\documentclass{ieeeaccess}
\usepackage{cite}
\usepackage{amsmath,amssymb,amsfonts}
\usepackage{algorithmic}
\usepackage{graphicx}
\usepackage{textcomp}
\usepackage{tabularx}

\usepackage{subcaption}

\usepackage{bm}
\makeatletter
\AtBeginDocument{\DeclareMathVersion{bold}
\SetSymbolFont{operators}{bold}{T1}{times}{b}{n}
\SetSymbolFont{NewLetters}{bold}{T1}{times}{b}{it}
\SetMathAlphabet{\mathrm}{bold}{T1}{times}{b}{n}
\SetMathAlphabet{\mathit}{bold}{T1}{times}{b}{it}
\SetMathAlphabet{\mathbf}{bold}{T1}{times}{b}{n}
\SetMathAlphabet{\mathtt}{bold}{OT1}{pcr}{b}{n}
\SetSymbolFont{symbols}{bold}{OMS}{cmsy}{b}{n}
\renewcommand\boldmath{\@nomath\boldmath\mathversion{bold}}}
\makeatother

\def\BibTeX{{\rm B\kern-.05em{\sc i\kern-.025em b}\kern-.08em
    T\kern-.1667em\lower.7ex\hbox{E}\kern-.125emX}}

\begin{document}
\history{Date of publication xxxx 00, 0000, date of current version xxxx 00, 0000.}
\history{This work has been submitted to the IEEE for possible publication. Copyright may be transferred without notice, after which this version may no longer be accessible.}
\doi{10.48550/arXiv.2401.07398}

\title{Cross Domain Early Crop Mapping using CropSTGAN}
\author{\uppercase{Yiqun WANG}\authorrefmark{1},
\uppercase{Hui HUANG}\authorrefmark{1}, and Radu STATE\authorrefmark{1}}

\address[1]{Services and Data Management Research Group (SEDAN), The Interdisciplinary Centre for Security, Reliability and Trust (SnT), University of Luxembourg, Luxembourg. (e-mail: yiqun.wang@uni.lu, hui.huang@uni.lu, radu.state@uni.lu)}

\markboth
{Author \headeretal: Preparation of Papers for IEEE TRANSACTIONS and JOURNALS}
{Author \headeretal: Preparation of Papers for IEEE TRANSACTIONS and JOURNALS}


\begin{abstract}
Driven by abundant satellite imagery, machine learning-based approaches have recently been promoted to generate high-resolution crop cultivation maps to support many agricultural applications. One of the major challenges faced by these approaches is the limited availability of ground truth labels. In the absence of ground truth, existing work usually adopts the "direct transfer strategy" that trains a classifier using historical labels collected from other regions and then applies the trained model to the target region. Unfortunately, the spectral features of crops exhibit inter-region and inter-annual variability due to changes in soil composition, climate conditions, and crop progress, the resultant models perform poorly on new and unseen regions or years.  Despite recent efforts, such as the application of the deep adaptation neural network (DANN) model structure in the deep adaptation crop classification network (DACCN), to tackle the above cross-domain challenges, their effectiveness diminishes significantly when there is a large dissimilarity between the source and target regions. This paper introduces the Crop Mapping Spectral-temporal Generative Adversarial Neural Network (CropSTGAN), a novel solution for cross-domain challenges, that doesn't require target domain labels. CropSTGAN learns to transform the target domain's spectral features to those of the source domain, effectively bridging large dissimilarities. Additionally, it employs an identity loss to maintain the intrinsic local structure of the data. Comprehensive experiments across various regions and years demonstrate the benefits and effectiveness of the proposed approach. In experiments, CropSTGAN is benchmarked against various state-of-the-art (SOTA) methods. Notably, CropSTGAN significantly outperforms these methods in scenarios with large data distribution dissimilarities between the target and source domains.
\end{abstract}

\begin{keywords}
Early Crop Mapping, Multispectral Image Data, Cross Domain, Domain Adaptation, CropSTGAN, Cropland Data Layer.
\end{keywords}

\titlepgskip=-21pt

\maketitle

\section{Introduction}
\PARstart{E}{arly} crop mapping, i.e. determining the cultivation regions of crops before their harvest season, is the fundamental building block for agricultural planning, resource allocation, crop insurance, risk management\cite{waldner2015mapping, singha2016land} and many other agricultural applications. Since the spectral features of vegetation are determined by their structure, leaf biochemistry and phonological stages \cite{xue2017significant}, time-series analysis on multi-spectral images captured by satellites is the dominant approach for land cover classifications. With the rapid growth in data volume and increasing accessibility of satellite imagery, deep learning (DL) based remote sensing has been promoted in recent years to produce high-resolution, high-accuracy crop cultivation maps \cite{joshi2023remote}. This class of approaches relies on a large amount of ground truth data, also known as labels, to train and validate the classification model. The ground truth can be obtained from surveys as first-hand labels\cite{zhang2019crop, you202110}, or use public datasets, such as the United States Department of Agriculture (USDA)’s Cropland Data Layer (CDL), as weak labels for model training \cite{boryan2011monitoring}. Different DL architectures, such as convolutional neural networks (CNNs), Temporal Convolutional Neural Network (TempCNN) \cite{pelletier2019temporal}, deep autoencoders and recurrent neural networks with long short-term memory (LSTM), are explored for crop mapping tasks \cite{wang2021new, hamidi2021auto, crisostomo2020rice}. The results suggest DL approaches outperform conventional support vector machine (SVM) and tree-based models in providing semantic information on the input images.

Unfortunately, obtaining appropriate ground truth for an arbitrary region is challenging. Public crop-type ground truth datasets, such as CDL, provide reliable and timely references for model training. However, these datasets are only available for a few countries and are usually released after harvest season. The collecting process can be costly, labour-intensive, and sometimes unfeasible, especially in underdeveloped countries. In the absence of ground truth, existing work usually adopts the "direct transfer strategy" that trains a classifier first using available labelled data for other regions and then applies the trained model to the target region \cite{hao2020transfer, ge2021transferable}. However, spectral features of crops have both inter-region variability and inter-annual variability due to changes in soil composition, climate conditions, and crop progress \cite{konduri2020mapping}. These variability collectively contribute to the distribution shift between the training data (source domain) and the test data (target domain), called the cross-domain issue. Consequently, the direct transfer strategy often leads to poor performance on new and unseen regions and years as it compromises an implicit assumption of the machine learning-based crop mapping approaches: the labelled training data from the source domain and the data from the target region are independent and identically distributed (i.i.d), or at least come from similar distributions. 


To address this cross-domain issue, including cross-region and cross-year issues, various methods have been developed aiming to enhance the model's ability on unseen domains. One effective approach involves training the model by incorporating multi-year data and the respective phonological metrics as the major inputs \cite{zhong2014efficient}. This method helps the model to capture the temporal variations in spectral patterns for target crops caused by changing environmental and climate conditions. Their methodology augments the model's proficiency in generalizing across diverse years within a singular region for which multi-annual ground truth data exists. However, the predicament of cross-regional issue remains unaddressed. 

To mitigate the challenges associated with both inter-regional and inter-annual cross-domain issues, methodologies such as Domain Adversarial Neural Networks (DANN) \cite{ajakan2014domain} and their derivatives, including Self-Training with Domain Adversarial Network (STDAN) \cite{kwak2022unsupervised}, Phenology Alignment Network (PAN) \cite{wang2021phenology}, and Deep Adaptation Crop Classification Network (DACCN) \cite{wang2023unsupervised}, have been deployed. These approaches endeavour to delineate invariant features from data across both target and source domains, leveraging these features for enhanced accuracy in crop classification mappings. However, these methodologies hinge on the presupposition that the data distributions between the target and source domains exhibit relatively minor disparities. Furthermore, all these methodologies operate under the assumption that the crop types are identical in both the source and target domains. In the STDAN study \cite{kwak2022unsupervised}, two case studies comprising four areas from Gangwon Province and Gyeongsang Province, Korea were selected. These areas, while not geographically distant, exhibit slight environmental differences. Similarly, the PAN study \cite{wang2021phenology} chose three areas from Sichuan Province, Hubei Province, and Anhui Province in China, all positioned at similar longitudes, also leading to minor environmental variations (Wang, 2021). The DACCN study \cite{wang2023unsupervised}, addressing cross-country issues for the first time in crop mapping experiments, selected sub-areas from Heilongjiang Province and Jilin Province, China as target domains, with four sub-areas from different states in the USA as source domains. This cross-country setup introduced significantly larger environmental variations compared to the previous studies. Additionally, the study included cross-year experiments in the USA. Notably, DACCN's performance metrics were significantly lower for cross-country issues than for cross-year experiments, indicating a reduction in efficiency due to substantial discrepancies in data distribution across countries. In other words, although these approaches adeptly tackle the issue of missing labels and diminish the detrimental impacts of domain shifts, their efficacy is constrained by relevantly substantial discrepancies in data distribution across the domains.

To address this domain shift challenge encountered in early crop mapping endeavors under relevantly substantial discrepancies in data distribution across the domains, this paper introduces the Crop Mapping Spectral-temporal Generative Adversarial Neural Network (CropSTGAN) framework. The system's primary goal is to identify a specified crop variety, such as corn, within a target area lacking labelled data. To fulfil this aim, the CropSTGAN framework employs an unsupervised domain adaptation (UDA) strategy, a technique for adapting a model from a source domain with labelled data to an unlabelled target domain. The innovative method learns a function that transforms the spectral characteristics features of the target area to the source domain while retaining their local structure. For instance, the spectral features of corn in the target domain may diverge from those in the source domain. The CropSTGAN framework first transforms the target domain's spectral features to resemble those of the source domain, thereby minimizing the discrepancies in feature patterns between the transformed target data and the original source data, all the while maintaining the distinguishability of corn and other land cover types within the target region. This process enables the straightforward application of a crop mapping model, trained within the source domain, to the target domain without diminishing its precision.

From the highest level, the proposed CropSTGAN system consists of three key components: the pre-processor, the CropSTGAN domain mapper, and the TempCNN crop mapper. The pre-processor module employs linear interpolation to fill gaps due to cloud coverage in the Multi-Spectral Images (MSI), ensuring a complete time series. The CropSTGAN domain mapper, a modification of the CycleGAN architecture \cite{zhu2017unpaired}, is designed with a specific structure to capture the temporal features from the time-series MSI data and transform time-series MSI data points from the target domain to the source domain. The CropSTGAN consists of two generator networks and two discriminator networks. The generators learn to transform data points from one domain to another domain, while the discriminators distinguish between the transformed data points provided by the generators and the original data points. Finally, the TempCNN crop mapper, a CNN-based model shared structure with the CropSTGAN discriminators, can be directly applied to the transformed target data to accurately determine the cultivation locations of the specified target crop on the target domain. 
To evaluate the distinct architecture of the CropSTGAN domain mapper, comparative experiments were undertaken utilizing a simpler, analogous structure named CropTGAN, alongside benchmarking against several state-of-the-art (SOTA) methodologies, including TempCNN and STDAN. The empirical findings corroborated that the CropSTGAN architecture enhances crop mapping accuracy under the relevantly substantial discrepancies in data distribution across the domains.

Our contributions can be summarized as follows:
\begin{itemize}
  \item {Propose the CropSTGAN framework, consisting of a pre-processor, a CropSTGAN domain mapper, and a TempCNN crop mapper, to address the cross-domain issue due to the inter-region and inter-year variations in remote sensing-based early crop mapping under substantial discrepancies in data distribution across the domains.}
  \item {Design the CropSTGAN domain mapper to capture the temporal and spectral features from the time-series MSI data and learn to transform the target domain data into the source domain.}
  \item {Conduct cross-region and cross-year experiments in the study areas from the USA and China to evaluate the CropSTGAN framework. The results demonstrate superior performance compared to CropTGAN and several SOTA methods, such as TempCNN and STDAN, confirming the effectiveness and accuracy of the CropSTGAN framework for cross-domain early crop mapping.}
  
\end{itemize}

The rest of this paper is organised as follows. Section \ref{sec:rw} describes related works. Section \ref{sec:ds} presents the data and study areas. The methodology is described in Section \ref{sec:md}. Experiment setup and results are presented in Section \ref{sec:er}. Finally, the discussion and conclusions are presented in Section \ref{sec:discussion}.

\section{Related Works}
\label{sec:rw}

Remote sensing-based crop-type mapping has undergone significant advancements with the adoption of machine learning methods. Traditionally, techniques such as SVM and random forest (RF) have been widely utilized for crop classification using remote sensing data \cite{zheng2015support, hao2015feature, saini2018crop}. However, the advent of deep learning has brought about a revolution in this field by leveraging their ability to automatically extract meaningful representations from data.

Deep learning models, such as LSTM networks and CNNs, have exhibited impressive performance in crop type mapping using remote sensing imagery \cite{wang2021new, crisostomo2020rice, sun2019using}. They excel at capturing temporal and spatial characteristics, enabling accurate classification of various crop types. By training on labelled crop datasets, these models can learn complex relationships between spectral, spatial, and temporal features, leading to improved accuracy in distinguishing crop types. 

To locate the target crop in the target domain without ground truth data, direct transfer techniques have been introduced, allowing the knowledge learned from a source domain, where labelled samples are abundant, to be directly applied to the target domain \cite{hao2020transfer, nowakowski2021crop}. It involves training models on regions with abundant labelled samples and directly applying them to other regions or years. This approach assumes that the knowledge learned from the source domain is applicable to the target domain. While direct transfer can be a simple and effective method, it may encounter a cross-domain problem when it comes to effectively capturing and comprehending the intricate relationships and variations in crop patterns across different regions and years.

To address the cross-year issue, training the model with multi-year crop data proves to be a valuable adaptation method. Recent studies, including \cite{zhong2014efficient}, emphasize the importance of incorporating multi-year crop data from a specific region as it enhances the model's understanding of temporal patterns, variations, and trends within that geographic context. This approach enhances the accuracy and reliability of crop classification by considering inter-annual climate variations and capturing the unique characteristics of the region. Leveraging region-specific multi-year data enables the model to become more robust to phenology shifts, ensuring consistent performance across different years. Training on multi-year crop data from the same region serves as an effective adaptation technique, resulting in more precise and dependable crop mapping outcomes within a particular geographic area. A key drawback of the multi-year crop data training method is its high reliance on labelled data. To effectively train deep learning models, a large number of accurately labelled samples are needed. However, collecting such a vast amount of labelled data from diverse regions and years is a challenging task. The process is time-consuming, costly, and often impractical due to the effort and resources required. Moreover, it fails to address the issue of the cross-region issue.

To tackle the challenges of cross-year and cross-region adaptability, current methods adopt two main strategies:

From a sample perspective, fine-tuning pre-trained models with a few high-quality target domain samples is a common practice. This method adapts the original model to the new data distribution. For example, \cite{tong2020land} improved deep models for nationwide land cover classification using high-confidence pseudo-labels. Similarly, \cite{hamrouni2021local} involved annotating new target domain samples to refine RF classifiers via active learning. However, this technique often necessitates labelling samples, which is not feasible for large-scale studies.

From a feature perspective, in the field of Unsupervised Domain Adaptation (UDA), earlier studies have developed DANN-based methods to align samples from various regions into a unified feature space, minimizing disparities in deep features. For instance, \cite{kwak2022unsupervised} introduced the STDAN, a novel unsupervised domain adaptation framework for crop type classification. Moreover, the DACCN \cite{wang2021phenology} and the PAN \cite{wang2023unsupervised}, extended the loss function using the Maximum Mean Discrepancy (MMD) and the Multiple Kernel variant of Maximum Mean Discrepancy (MK-MMD), achieving improved accuracy compared to CNN and LSTM methods without domain adaptation. These methods often presuppose minor difference between the target and source domain data distributions. However, when these differences are substantial, the effectiveness diminishes.

This paper introduces the CropSTGAN framework, designed to tackle the challenge of cross-domain early crop mapping without the need for labelled data from the target domain, even when there are significant differences between the data distributions of the target and source domains. Unlike DANN-variant methods that focus on extracting invariant features, CropSTGAN transforms unlabeled target domain data into the source domain. The CropSTGAN system comprises three main components: a pre-processor, a CropSTGAN domain mapper, and a TempCNN crop mapper. The domain mapper, a key part of CropSTGAN, transforms time-series MSI data prepared by the pre-processor from the target domain into the source domain. Subsequently, a TempCNN crop mapper, pre-trained with labelled data from the source domain, processes this transformed data to produce the crop mapping results for the target domain. As far as our knowledge extends, we are the pioneering contributors to tackling the cross-domain issue in early crop mapping through the utilization of a model based on GAN combined with specially designed identity losses.

\section{Data and Study areas}
\label{sec:ds}

    \subsection{Study area}
    \label{sec:sa}
    
    \begin{figure}[h]
    \centering
    \subfloat[]{\includegraphics[width=1.5in,]{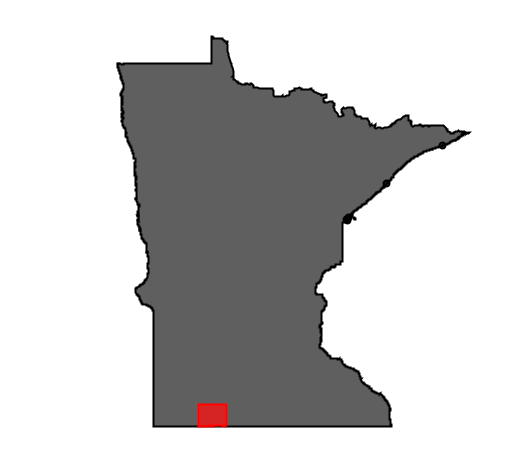}}%
    \label{fig:USA_geo}
    \hfil
    \subfloat[]{\includegraphics[width=1.5in,]{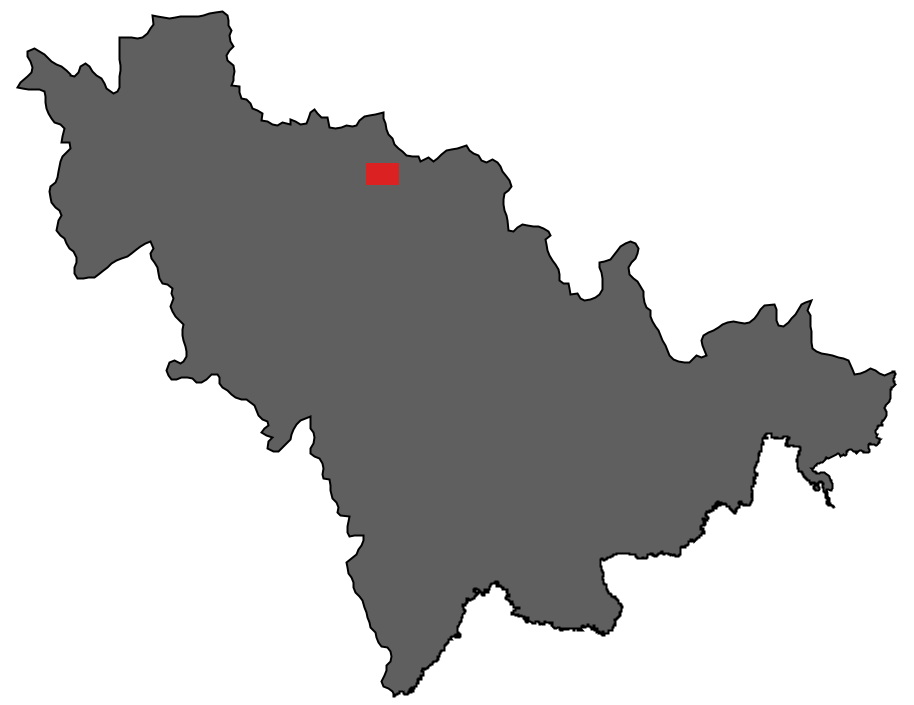}}%
    \label{fig:China_geo}  
    \hfill
    \caption{The Study Areas. (a) Jackson County, Minnesota, USA. (b) The Study Area (from 44.97$^\circ$N to 45.14$^\circ$N latitude and from 125.49$^\circ$W to 125.86$^\circ$W longitude), Jilin Province, China.}
    \label{fig:domains}
    \end{figure}

 \begin{figure}[]
    \centering
    \subfloat[ ]
    {\includegraphics[width=3in]{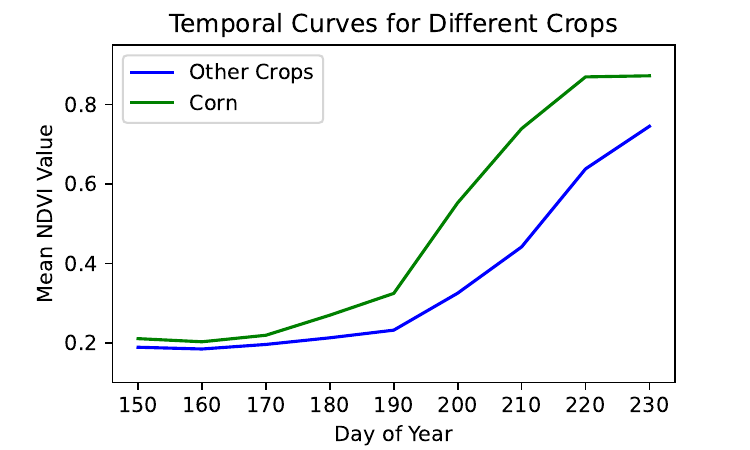}}%
    \label{fig:Jackson_NDVI_2019}
    \hfil
    \subfloat[ ]{\includegraphics[width=3in]{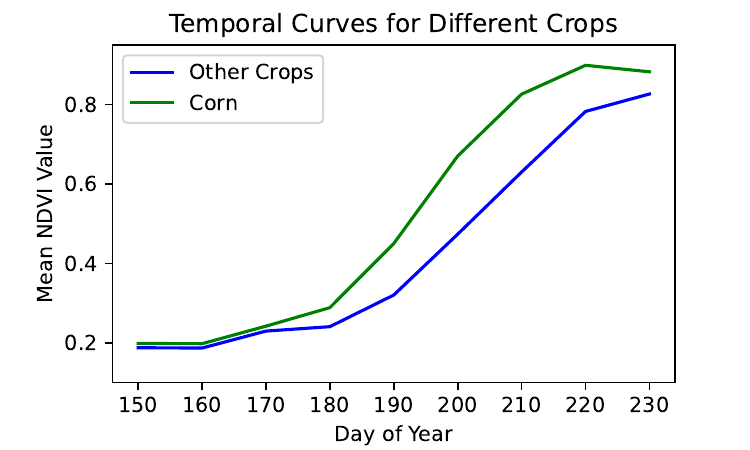}}%
    \label{fig:Jackson_NDVI_2020}
    \hfil
    \subfloat[ ]{\includegraphics[width=3in]{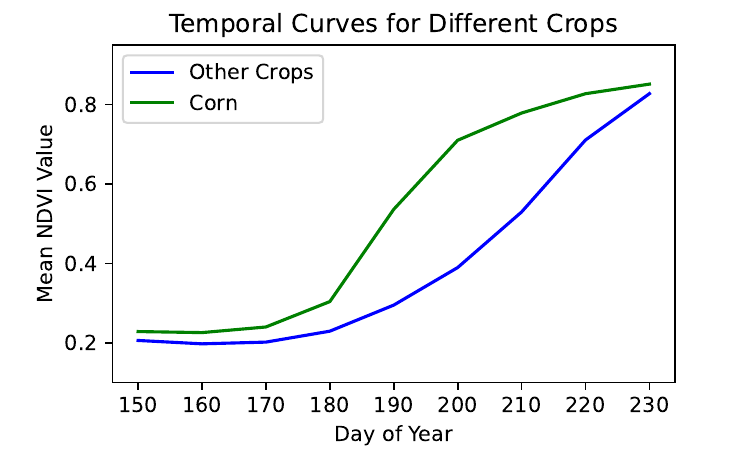}}%
    \label{fig:Jackson_NDVI_2021}
    \hfil
    \subfloat[ ]{\includegraphics[width=3in]{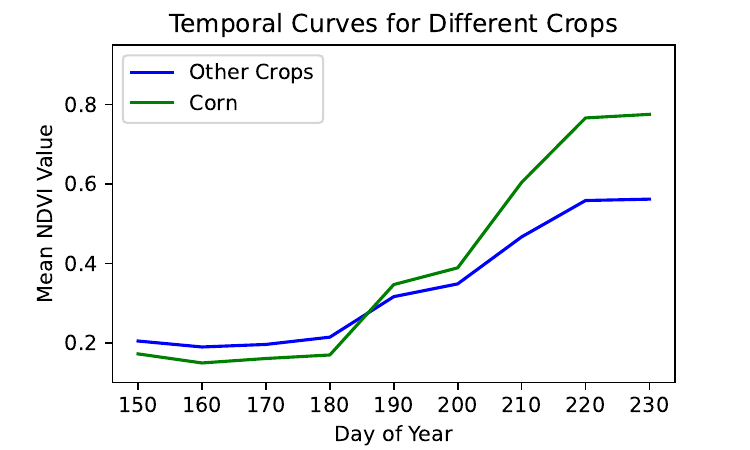}}%
    \label{fig:NDVI_China}
    \hfil
    \caption{The Average Time-series NDVI of Corn for the Source Domain and Target Domain. (a) Jackson County, 2019. (b) Jackson County, 2020. (c) Jackson County, 2021. (d) The study area of China, 2019.}
    \label{fig:ndvi_curve}
\end{figure}

This paper chose corn as the example crop to demonstrate the cross-domain capability of the proposed approach. The objective is to map corn cultivation locations in a target domain using models trained by labelled data from a source domain. The target and source domains could differ in their geographic locations (i.e., cross-region) or be in the same region in different years (i.e., cross-year). Note that, the proposed approach can be applied to other types of crops. Toward this end, the study areas include the two regions: Jackon County of the USA, and one study area from Jilin Province, China. For a visual depiction of the geographical locations of these areas, please refer to Figure \ref{fig:domains}.

    \begin{table}[]
    \centering
    \caption{The environmental conditions of the source domain and the target domains. "SA" represents the Study Area. "T" represents the yearly average temperature. "P" stands for the average hourly precipitation. "E" signifies the average hourly evaporation. "R" indicates the surface net solar radiation.}
    \label{tab:environmental}
    \begin{tabular}{|c|c|c|c|c|c|}
    \hline
       SA & Year & \textbf{T}(K) & \textbf{P}(mm/h) & \textbf{E}(mm/h) & \textbf{R}(kJ/m\textasciicircum{}2) \\\hline
        Jackson & 2019 & 289.02          & 2.60                 & -1.01             & 4688.12      \\
        Jackson & 2020 & 289.93          & 1.48                 & -1.11             & 5118.46                                         \\
        Jackson & 2021 & 290.76          & 1.06                 & -1.06             & 5245.53                                             \\
     
        SA in China & 2019   & 290.03          & 1.99                & -1.99             & 12542.54       \\\hline
    \end{tabular}
    \end{table}
     
     The study areas are characterized by unique environmental conditions such as temperature, precipitation, elevation, and solar radiation, which also vary over years. Table \ref{tab:environmental} presents the environmental metrics for these areas. Additionally, the corn cropping schedules differ among the regions, as depicted in Figure \ref{fig:cropcalendar}. These variations influence corn growth, resulting in distinct MSI feature patterns of the same crop across the different areas and time periods. Figures 2(a), 2(b), and 2(c) show the average NDVI values curves for the early growth stages of corn and other crops in Jackson County from 2019 to 2021. Figure 2(d) presents the curves in the study area of China. The corn growth NDVI patterns in Jackson County vary across the different years. Additionally, a clear distinction is evident between the NDVI values in Jackson County and those in the Chinese study area.
     

    \subsection{Reference Data}
    \label{sec:ra}
    
    The CDL is used as the ground truth for the source domain. The CDL \cite{boryan2011monitoring}, a crop-specific land cover raster map dataset available for the entire conterminous U.S. land area (CONUS) at 30 m resolution provided by the USDA, regularly provides information on the annual temporal and spatial distribution of corn, as well as the area dedicated to its cultivation. However, for the target domain, official ground reference data for China is currently unavailable. Fortunately, \cite{you202110} published maps of corn and soybean with a spatial resolution of 10 meters for Northeast China from 2017 to 2019. As a result, the 2019 crop maps from \cite{you202110} are regarded as the ground truth reference for the study area in China. 
    


    \subsection{Remote Sensing Data}
    \label{sec:si}
    
    The remote sensing data in this work are MSI images captured by the Sentinel-2 satellites, which have been widely used for many agricultural applications in the community \cite{wang2021new, blickensdorfer2022mapping}. Sentinel-2 provides high-resolution MSI images (up to 10m) with a revisit time of 5 days, allowing for frequent monitoring of crop growth and changes. Its wide spectral coverage, including visible, near-infrared, and shortwave infrared bands, enables accurate assessment of vegetation health, crop type identification, and mapping. In this study, 6 bands, including B2 (Blue), B3 (Green), B4 (Red), B8 (Near-Infrared), B11 (Shortwave Infrared 1) and B12 (Shortwave Infrared 2), are used to map the target crop in early seasons. 

\begin{figure}[]
        \centering
        \includegraphics[width=3in]{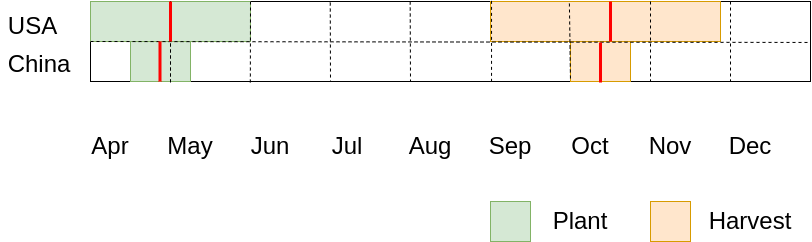}
        \caption{The Crop Calendar for Corn. The red lines represent the average dates.}
        \label{fig:cropcalendar}
    \end{figure}

    As described, our primary objective is to locate the specific crop (corn) with cross-domain problems at an earlier growth stage. As a result, the time series remote sensing data should start after the planting period and conclude before the onset of the crop harvest period. Referring to Figure \ref{fig:cropcalendar}, it can be observed that the corn harvesting season typically starts in early September in the USA and begins in early October in China. Planting, on the other hand, starts at the beginning of April in the USA, and in mid-April in Jilin Province, China. As a result, the date of our annual collection of remote sensing data spans from May 1st to July 30th, encompassing a total of nine-time points with a ten-day observation window. 
    

    Additionally, this work employs the Dynamic World dataset \cite{brown2022dynamic}. It is a high-resolution 10m near-real-time (NRT) Land Use/Land Cover (LULC) dataset and features class probabilities and label data for nine distinct categories, including cropland. It is utilized to identify and select areas of cropland, enabling us to maintain a concentrated analysis solely within these regions.
    
\section{Methodology}
\label{sec:md}

    \subsection{Problem Statement}
    Let $\mathbf{X}$ denote the time series remote sensing input data and $\mathbf{Y}$ denote the ground truth labels. Each sample $\mathbf{x}$ can be expressed as a temporal form $[\mathbf{x}_{1}, \mathbf{x}_2, ..., \mathbf{x}_{t}]$, where $\mathbf{x}_{i}$ represents input at time $i$. $\mathbf{x}_i$ can be further expanded as $[\mathbf{x}_{i1}, \mathbf{x}_{i2}, ..., \mathbf{x}_{ib}]$, containing multi-spectral bands information from band $1$ to band $b$. Each ground truth label, denoted as $y$, is represented as a binary number, which corresponds to the categories of corn and other crops. Let $\mathbf{X}_{t}$ denote the time series input target data, $\mathbf{X}_{s}$ denote the time series input source data, and $\mathbf{Y}_{s}$ denote the GT for $\mathbf{X}_{s}$. Each sample $\mathbf{x}_s$ has a corresponding $\mathbf{y}_s$. 
    
    Our objective is to identify target crops to get target labels $\mathbf{Y}_{t}$ in the target domain during their early growth stages, utilizing labelled source domain data ($\mathbf{X}_{s}$, $\mathbf{Y}_{s}$) and unlabelled target domain data ($\mathbf{X}_{t}$). This approach addresses challenges related to the cross-domain issue and the large data distribution discrepancies issue.

    \subsection{System Overview}
        \label{sec:so}

    \begin{figure*}[!t]
        \centering
        \includegraphics[width=6in]{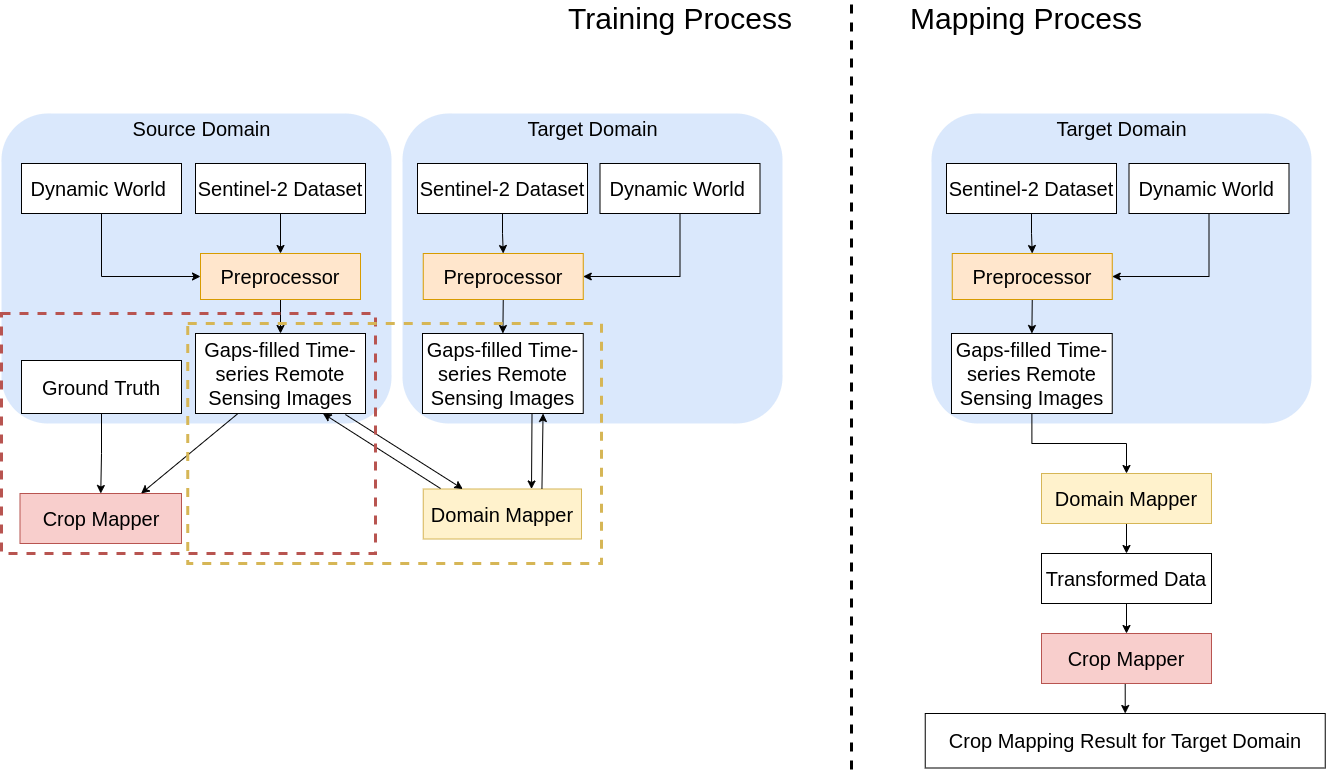}
        \caption{The Architecture of the CropSTGAN Framework.}
        \label{fig:design}
    \end{figure*}

        
    The architecture of our CropSTGAN framework is depicted in Figure \ref{fig:design}, showcasing both the training and mapping dataflows. This system includes three main components: the pre-processor, the CropSTGAN domain mapper, and the TempCNN crop mapper. During training, the pre-processor enhances MSI images quality by filling gaps through linear interpolation, ensuring complete MSI image series $\mathbf{X}_{s}$ and $\mathbf{X}_{t}$ for both source and target domains. The CropSTGAN domain mapper, trained with randomly sampled data from these MSI images, enables the transfer of time-series remote sensing data between domains: $\mathbf{X}_{t} \rightarrow \mathbf{X}_{s}$ and $\mathbf{X}_{s} \rightarrow \mathbf{X}_{t}$. The TempCNN crop mapper is then trained on labelled source domain data ($\mathbf{X}_{s}$, $\mathbf{Y}_{s}$) to identify the target crop in the source domain at an early stage.
    
    For mapping, the process begins with the pre-processor supplying the target domain MSI image series $\mathbf{X}_{t}$. These images are then transformed by the CropSTGAN domain mapper from the target to the source domain as $\mathbf{X}_{s}^{'}$. The trained TempCNN crop mapper uses these transformed images $\mathbf{X}_{s}^{'}$ to locate the target crop $\mathbf{Y}_{t}$ in the target domain. This system design ensures accurate and reliable early-stage crop mapping in the target domain, particularly useful when labelled data is scarce or nonexistent, thereby overcoming the challenges of cross-domain mapping.
    
    \subsection{Pre-processor}
    \label{sec:gf}

        \begin{figure}[!t]
            \centering
            \includegraphics[width=2.5in]{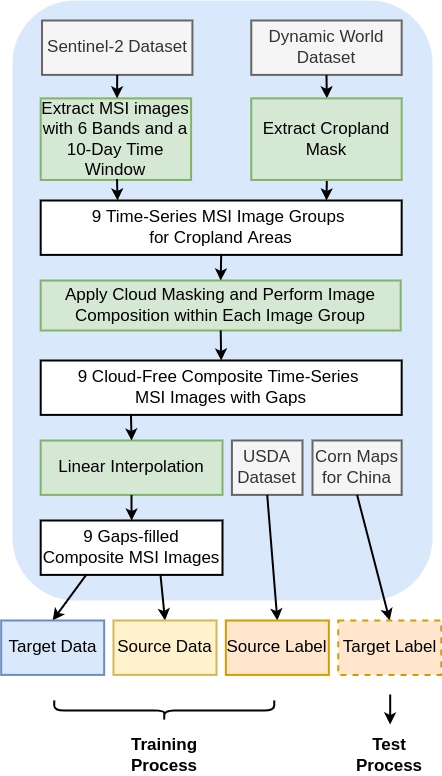}
            \caption{The Architecture of the Pre-processor.}
            \label{fig:pre-processor}
        \end{figure}

    The pre-processor aims at providing complete time-series MSI data by filling gaps between MSI images due to cloud cover, atmospheric interference, or sensor limitations. The processing pipeline is outlined in Figure \ref{fig:pre-processor}. 

    The procedure illustrated in the figure begins with the collection of MSI images covering the entire targeted study areas, sourced from the Sentinel-2 Dataset at consistent 10-day intervals. The primary objective is to detect the target crop at an early growth stage, addressing cross-domain challenges. Therefore, the selected time series of remote sensing data spans from May 1st to July 30th, starting post-planting and concluding pre-harvest, resulting in nine image sets. To refine the focus on agricultural lands, the method employs a cropland mask from the Dynamic World dataset during the crop's growing season, adjusting it to a uniform 30-meter resolution. This step ensures the exclusion of non-agricultural areas, concentrating the analysis on crop areas within the nine groups of MSI images. For each set, cloud detection is performed using Sentinel Hub's tool, filtering out cloud-covered areas. The process averages the values of cloud-free images to produce a composite MSI image with a clear 30-meter resolution. This approach yields nine high-quality, cloud-free time-series MSI images, each comprising six bands and maintaining a 30-meter spatial resolution.

    \subsection{CropSTGAN Domain Mapper}
    \label{sec:dm}
    
        \begin{figure}[!t]
        \centering
        
        \subfloat[]{\includegraphics[width=3.3in]{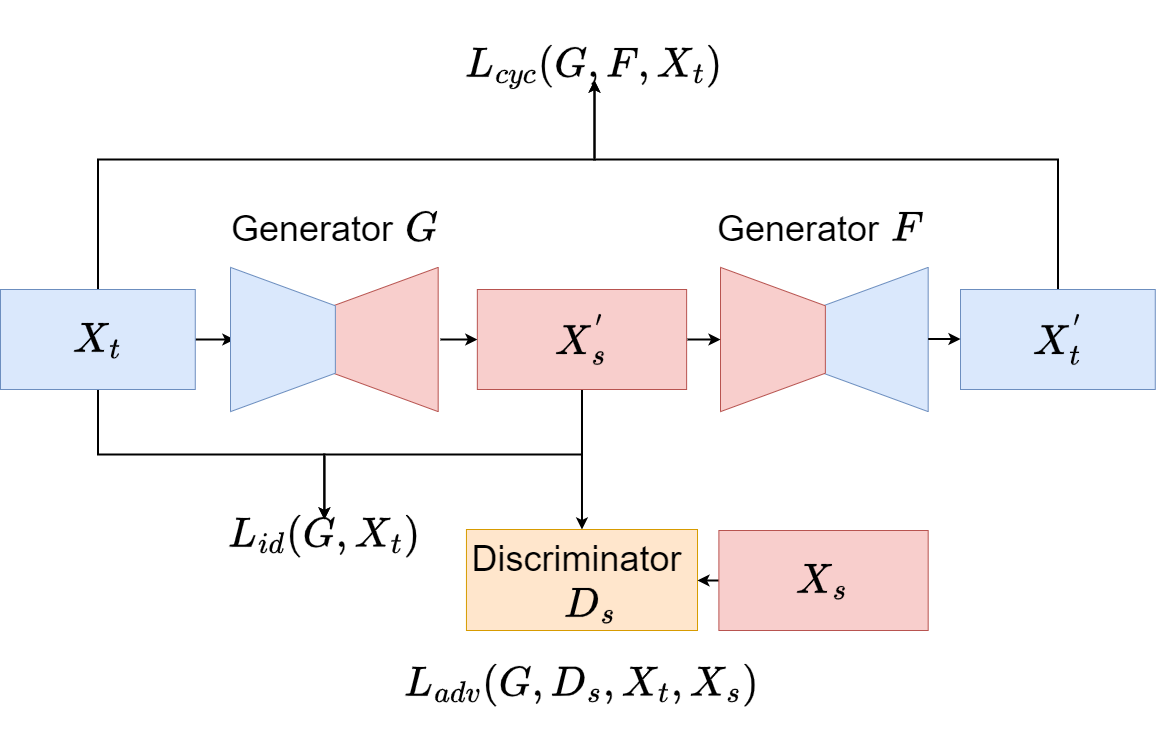}}%
        \label{fig:cropGAN_G}
        \hfil
        \subfloat[]{\includegraphics[width=3.3in]{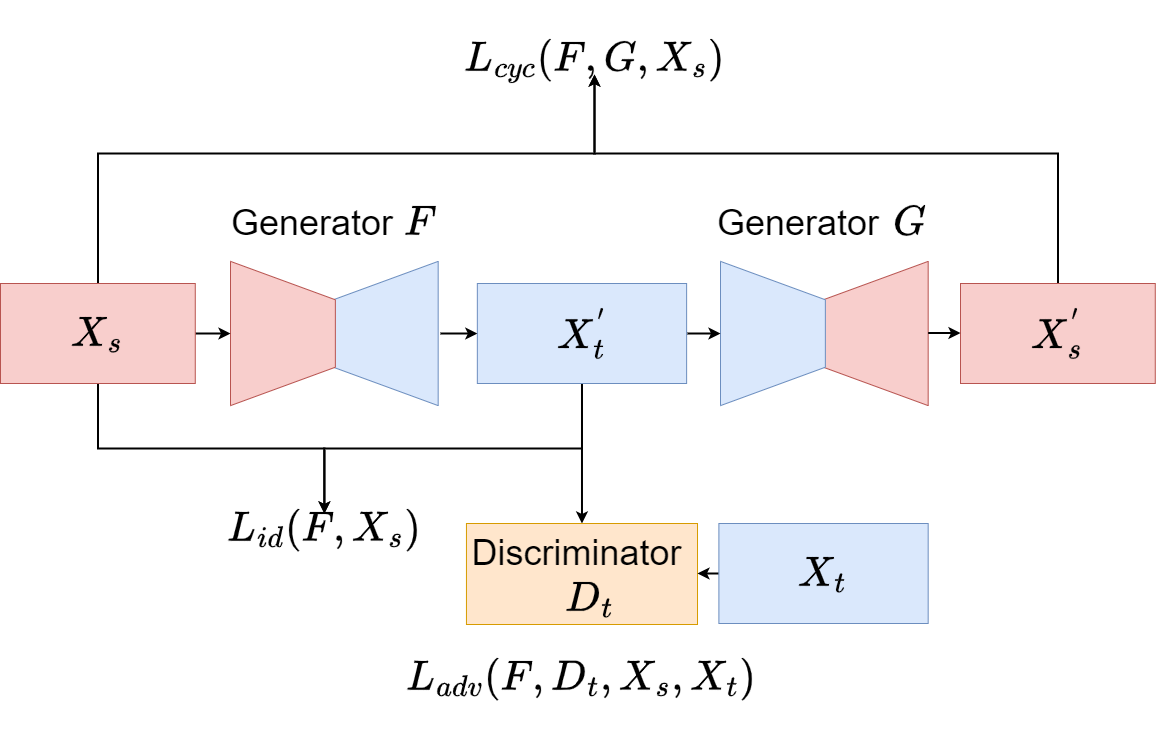}}%
        \label{fig:cropGAN_F}
        
        \caption{The CropSTGAN Domain Mapper Structure. (a) Cycle 1: The training process involves data flowing from the target domain to the source domain and then back to the target domain. The notation $\mathbf{X}_{s}^{'}$ represents the transformed target domain data, while $\mathbf{X}_{t}^{'}$ denotes the data transformed back to the target domain. (b) Cycle 2: The training process involves data flowing from the source domain to the target domain and then back to the source domain. The notation $\mathbf{X}_{t}^{'}$ represents the transformed source domain data, while $\mathbf{X}_{s}^{'}$ denotes the data transformed back to the source domain.}
        \label{fig:cropGAN}
        \end{figure}
    
    \subsubsection{The Generators and Discriminators Structures}

            \begin{table}[t!]
        \caption{The Generator Structure. Each encoder layer consists of an instance normalization layer, an activation layer and a pooling layer. Each decoder layer consists of an instance normalization layer, an activation layer and an upsampling layer.}
        \centering
        \begin{tabular}{|c| c| c| c|}
        \hline
        \textbf{Layer} & \textbf{Input Shape} & \textbf{Output Shape} & \textbf{Activation} \\
        \hline
        Input & 9$\times$6 & 9$\times$6 & - \\
        \hline
        Encoder 1 & 9$\times$6 & 9$\times$1$\times$4 & LeakyReLU \\ 
        Encoder 2 & 9$\times$1$\times$4 & 7$\times$1$\times$8 & LeakyReLU \\
        Encoder 3 & 7$\times$1$\times$8 & 5$\times$1$\times$16 & LeakyReLU \\
        Encoder 4 & 5$\times$1$\times$16 & 3$\times$1$\times$32 & LeakyReLU \\
        \hline
        Decoder 4 & 3$\times$1$\times$32 & 5$\times$1$\times$16 & LeakyReLU \\
        Decoder 3 & 5$\times$1$\times$16 & 7$\times$1$\times$8 & LeakyReLU \\
        Decoder 2 & 7$\times$1$\times$8 & 9$\times$1$\times$4 & LeakyReLU \\
        Decoder 1 & 9$\times$1$\times$4 & 9$\times$6$\times$4 & LeakyReLU \\
        \hline
        Output & 9$\times$6$\times$4 & 9$\times$6 & ReLU \\
        \hline

        \end{tabular}

        \label{tab:generator}
        \end{table}

        \begin{table}[t!]
        \caption{The Discriminator/Crop Mapper Structure. The last layer of the Discriminator uses ReLU as its activation function, while the Classifier employs Softmax.}
        \centering
        \begin{tabular}{|c| c| c| c|}
        \hline
        \textbf{Layer} & \textbf{Input Shape} & \textbf{Output Shape} & \textbf{Activation} \\
        \hline
        Input & 9$\times$6 & 9$\times$6 & - \\
        \hline
        Conv 1 & 9$\times$6 & 9$\times$1$\times$8 & LeakyReLU \\
        Conv 2 & 9$\times$1$\times$8 & 7$\times$1$\times$8 & LeakyReLU \\
        Conv 3 & 7$\times$1$\times$8 & 5$\times$1$\times$8 & LeakyReLU \\
        Conv 4 & 5$\times$1$\times$8 & 3$\times$1$\times$8 & LeakyReLU \\ 
        Flatten & 3$\times$1$\times$8 & 24 & - \\
        FC1    & 24    & 4  & ReLU\\
        \hline
        Output & 4 & 1 & ReLU or Softmax \\
        \hline
        \end{tabular}
        \label{tab:discriminator}
        \end{table}
        
        The domain mapper is referred as CropSTGAN, which stands for Crop Mapping Spectral-temporal Generative Adversarial Network. CropSTGAN domain mapper serves as the key component in our framework for transforming time-series remote sensing data points between different domains. As shown in Figure \ref{fig:cropGAN}, the domain mapper consists of two generator networks, $G$ and $F$, and two discriminator networks, $D_{X}$ and $D_{Y}$. In the training process, the generators and discriminators play distinct roles. The generators' primary task is to transform the data from one domain to another: $G: \mathbf{X}_{t} \rightarrow \mathbf{X}_{s}^{'}$ and $F: \mathbf{X}_{s} \rightarrow \mathbf{X}_{t}^{'}$, while the discriminator's main objective is to differentiate between real and transformed data. As the training progresses, both networks improve iteratively until an equilibrium is reached where the generator generates highly realistic data and the discriminator cannot reliably distinguish between real and fake samples. For example, when the discriminator $D_{s}$ can not distinguish whether the transformed data $\mathbf{X}_{s}^{'}$ generated by the generator $G$ is from the source domain or not, it means that the generator $G$ has been well-trained.  

        The generator and discriminator structures are depicted in Table \ref{tab:generator} and Table \ref{tab:discriminator}, respectively. Each generator consists of four encoders and four decoders. Its first encoder layer, equipped with $4$ filters with a $3$ kernel size, captures temporal features across all 6 spectral bands. Subsequent encoders use the same kernel size but increase filter counts to 8, 16, and 32, respectively. The decoders, designed to mirror the encoders, reverse the encoding process, thereby reconstructing the original input features. 
        
        The discriminator shares a similar design with the generator's encoder, with the distinction that its convolution layers uniformly utilize 8 filters. It then incorporates a flatten layer and a fully connected layer, with dimensions of 24 and 4, respectively. Its final layer is a fully connected layer with a single output, aimed at distinguishing between the transformed target domain data and the source domain data.

        \begin{table}[]
        \caption{The Generator Structure of CropTGAN. Each encoder layer consists of an instance normalization layer, an activation layer and a pooling layer. Each decoder layer consists of an instance normalization layer, an activation layer and an upsampling layer.}
        \centering
        \begin{tabular}{|c| c| c| c|}
        \hline
        \textbf{Layer} & \textbf{Input Shape} & \textbf{Output Shape} & \textbf{Activation} \\
        \hline
        Input & 9$\times$6 & 9$\times$6 & - \\
        \hline
        Encoder 1 & 9$\times$6 & 9$\times$6$\times$4 & LeakyReLU \\ 
        Encoder 2 & 9$\times$6$\times$4 & 7$\times$6$\times$8 & LeakyReLU \\
        Encoder 3 & 7$\times$6$\times$8 & 5$\times$6$\times$16 & LeakyReLU \\
        Encoder 4 & 5$\times$6$\times$16 & 3$\times$6$\times$32 & LeakyReLU \\
        \hline
        Decoder 4 & 3$\times$6$\times$32 & 5$\times$6$\times$16 & LeakyReLU \\
        Decoder 3 & 5$\times$6$\times$16 & 7$\times$6$\times$8 & LeakyReLU \\
        Decoder 2 & 7$\times$6$\times$8 & 9$\times$6$\times$4 & LeakyReLU \\
        Decoder 1 & 9$\times$6$\times$4 & 9$\times$6$\times$4 & LeakyReLU \\
        \hline
        Output & 9$\times$6$\times$4 & 9$\times$6 & ReLU \\
        \hline

        \end{tabular}

        \label{tab:generator_CropTGAN}
        \end{table}

        \begin{table}[]
        \caption{The Discriminator/Crop Mapper Structure of CropTGAN. The last layer of the Discriminator uses ReLU as its activation function, while the Classifier employs Softmax.}
        \centering
        \begin{tabular}{|c| c| c| c|}
        \hline
        \textbf{Layer} & \textbf{Input Shape} & \textbf{Output Shape} & \textbf{Activation} \\
        \hline
        Input & 9$\times$6 & 9$\times$6 & - \\
        \hline
        Conv 1 & 9$\times$6 & 9$\times$6$\times$8 & LeakyReLU \\
        Conv 2 & 9$\times$6$\times$8 & 7$\times$6$\times$8 & LeakyReLU \\
        Conv 3 & 7$\times$6$\times$8 & 5$\times$6$\times$8 & LeakyReLU \\
        Conv 4 & 5$\times$6$\times$8 & 3$\times$6$\times$8 & LeakyReLU \\ 
        Flatten & 3$\times$6$\times$8 & 144 & - \\
        FC1    & 144    & 4  & ReLU\\
        \hline
        Output & 4 & 1 & ReLU or Softmax \\
        \hline
        \end{tabular}
        \label{tab:discriminator_CropTGAN}
        \end{table}

        To assess the distinctive architecture of the CropSTGAN domain mapper, we conducted comparative experiments with a simpler yet analogous model named CropTGAN. This model features encoders with identical filter counts and sizes as those in CropSTGAN. The primary distinction between the generators and discriminators of the two models lies in the configuration of the first layers. Specifically, the CropTGAN generator and discriminator independently extract features from the temporal dimension on each band in their respective first encoder and convolutional layers. The details of the generators and discriminators are documented in Table \ref{tab:generator_CropTGAN} and Table \ref{tab:discriminator_CropTGAN}.

        \subsubsection{Losses}

        As shown in Figure \ref{fig:cropGAN}, there are two cycle training processes:
        \begin{itemize}
            \item In the first cycle training process (Cycle 1), data from the target domain is first transformed to the source domain and then reverted back to the target domain: $G: \mathbf{X}_{t} \rightarrow \mathbf{X}_{s}^{'} \rightarrow \mathbf{X}_{t}^{'}$ 
            and 
            \item  In the second cycle training process (Cycle 2), data from the source domain is first transformed to the target domain and then reverted back to the source domain: $F: \mathbf{X}_{s} \rightarrow \mathbf{X}_{t}^{'} \rightarrow \mathbf{X}_{s}^{'}$
        \end{itemize}
        
        To ensure the creation of effective generators, the losses for both training processes are established as follows:

        \subsubsection*{\bf Adversarial Loss} This loss function is inspired by GANs and encourages the generator to produce transformed data from one domain that is indistinguishable from real data in the other domain. It is computed by the discriminator network, which aims to classify the generated data as fake while the generator aims to fool the discriminator by generating realistic data (source domain data).

        \begin{equation}
                \begin{split}
                   \mathcal{L}_{\text{adv1}}(G, D_s, \mathbf{X}_{t}, \mathbf{X}_{s}) =  & \mathbb{E}_{\mathbf{x}_{s} \sim p_{\text{data}}(\mathbf{x}_{s})}[\log D_s(\mathbf{x}_{s})] + \\
                    &  \mathbb{E}_{\mathbf{x}_{t} \sim p_{\text{data}}(\mathbf{x}_{t})}[1 - \log D_s(G(\mathbf{x}_{t}))]
                \end{split}
                \label{eq:AL}
        \end{equation}

        \begin{equation}
                \begin{split}
                   \mathcal{L}_{\text{adv2}}(F, D_t, \mathbf{X}_{s}, \mathbf{X}_{t}) =  & \mathbb{E}_{\mathbf{x}_{t} \sim p_{\text{data}}(\mathbf{x}_{t})}[\log D_t(\mathbf{x}_{t})] + \\
                    &  \mathbb{E}_{\mathbf{x}_{s} \sim p_{\text{data}}(\mathbf{x}_{s})}[1 - \log D_t(G(\mathbf{x}_{s}))]
                \end{split}
                \label{eq:AL}
        \end{equation}



        \subsubsection*{\bf Cycle Consistency Loss} The cycle consistency loss ensures that the transformed data from one domain to another domain and back to the original domain is consistent with the original data. It measures the difference between the original input data and the data reconstructed after going through both generator mappings. Through the minimization of this loss, the preservation of crucial features during the transformation process is enforced.

        \begin{equation}
        \mathcal{L}_{\text{cyc1}}(G, F, \mathbf{X}_{t}) = \mathbb{E}_{\mathbf{x}_{t} \sim p_{\text{data}}(\mathbf{x}_{t})}[|| \mathbf{x}_{t} - F(G(\mathbf{x}_{t})) ||_1]
        \end{equation}
        
        \begin{equation}
        \mathcal{L}_{\text{cyc2}}(F, G, \mathbf{X}_{s}) = \mathbb{E}_{\mathbf{x}_{s} \sim p_{\text{data}}(\mathbf{x}_{s})}[|| \mathbf{x}_{s} - G(F(\mathbf{x}_{s})) ||_1]
        \end{equation}

        \subsubsection*{\bf Identity Loss} The identity loss encourages the generator to preserve the identity of the input data. It computes the difference between the generator output and the input data. The objective of minimizing this loss is to ensure that the generator does not make unnecessary alterations to the data and maintains its essential characteristics.

        \begin{equation}
        \mathcal{L}_{\text{id1}}(G, \mathbf{X}_{t}) = \mathbb{E}_{\mathbf{x}_{t} \sim p_{\text{data}}(\mathbf{x}_{t})}[|| G(\mathbf{x}_{t}) - \mathbf{x}_{t} ||_1]
        \end{equation}
        
        \begin{equation}
        \mathcal{L}_{\text{id2}}(F, \mathbf{X}_{s}) = \mathbb{E}_{\mathbf{x}_{s} \sim p_{\text{data}}(\mathbf{x}_{s})}[|| F(\mathbf{x}_{s}) - \mathbf{x}_{s} ||_1]
        \end{equation}

        \subsubsection*{\bf Total Loss} The total loss is a combination of these losses, weighted by respective coefficients:
        \begin{equation}
           \begin{split}
                \mathcal{L}_{\text{total}} = &   \alpha (\mathcal{L}_{\text{adv1}}(G, D_s, \mathbf{X}_{t}, \mathbf{X}_{s}) + \mathcal{L}_{\text{adv2}}(F, D_t, \mathbf{X}_{s}, \mathbf{X}_{t})) + \\
                & \beta(\mathcal{L}_{\text{cyc1}}(G, F, \mathbf{X}_{t}) +  \mathcal{L}_{\text{cyc2}}(F, G, \mathbf{X}_{s})) + \\
                & \sigma(\mathcal{L}_{\text{id1}}(G, \mathbf{X}_{t}) + \mathcal{L}_{\text{id2}}(F, \mathbf{X}_{s}))
            \end{split}
        \end{equation},
        where $\alpha, \beta, \sigma$ are the weight parameters. The total loss can be expressed as a minimax function: 
        \begin{equation}
        \min_{G, F} \max_{D_t, D_s} \mathcal{L}_{\text{total}}(G, F, D_t, D_s, \mathbf{X}_{t}, \mathbf{X}_{s})
        \end{equation},
        where the generator seeks to minimize the loss while the discriminator aims to maximize it to achieve high-quality time-series MSI data transformation while maintaining consistency and identity preservation.

        \subsubsection*{\bf Training stop criteria based on the Total Loss}


        
        
        The total loss is employed as an evaluation metric. After the GropSTGAN network has reached a stable state for the training process after several training epochs, the model is selected associated with the smallest total loss. This particular model is then designated as our well-trained model. This approach allows us to tackle the absence of paired datasets for evaluation in CropSTGAN and obtain an objective measure of the model's performance.
       
    \subsection{TempCNN Crop Mapper}
    \label{sec:cm}        
        
        The TempCNN crop mapper denoted as $C$, structured similarly to the CropSTGAN domain mapper's discriminators $D_x$ and $D_y$ as outlined in Table \ref{tab:discriminator}, is trained solely with labelled source domain data ($\mathbf{X}_{s}$, $\mathbf{Y}_{s}$). The binary cross-entropy loss:
        \begin{equation}
            \mathcal{L}_{\text{class}}(C) = -\sum_{s} \left( y_s \log(C(\mathbf{x}_s)) + (1 - y_s) \log(1 - C(\mathbf{x}_s)) \right)
        \end{equation}
        , is used as the training loss for the TempCNN crop mapper. Its purpose is to locate the target crop within the source domain. However, applying this mapper directly in the target domain often leads to subpar results due to discrepancies in the variation of data distribution between the domains. To address this issue, the target domain's MSI data must first be transformed to the source domain using the domain mapper's generator. As illustrated in the right portion of Figure \ref{fig:design}, this transformed data is then processed by the TempCNN crop mapper to generate mapping results:
        \begin{equation}
            \mathbf{Y}_{t} = C(G(\mathbf{X}_{t}))
        \end{equation}
        This integrated approach ensures accurate and reliable early crop mapping in the target domain, even in the absence of the target domain ground truth data.

\section{Experiment setup and Results}
\label{sec:er}

            \begin{table*}[t!]
                \centering
                \caption{Cross-Year Experiment Metrics Comparison: Targeting Jackson County 2020. The best metrics are indicated in bold, while the second-best metrics of the baseline methods are underlined.}
                   
                    \begin{tabular}{|c|ccc|ccc|ccc|ccc|}
                    \hline
                    & \multicolumn{3}{c|}{\textbf{CropSTGAN}} & \multicolumn{3}{c|}{\textbf{CropTGAN}} & \multicolumn{3}{c|}{\textbf{STDAN}} & \multicolumn{3}{c|}{\textbf{TempCNN}} \\ 
                    \hline
                    &OA  & F1  &  Kappa&OA  & F1  &  Kappa  & OA  & F1  &  Kappa & OA   & F1  &  Kappa           \\ 
                    S1&\underline{0.9300} & \underline{0.9311}&\underline{0.8596}&0.9167& 0.9142&0.8339& \bf{0.9389} &\bf{0.9386} & \bf{0.8769} &0.6162 &0.4304 & 0.2623 \\ 
                    S2&0.9220 &0.9218 &0.8441&0.8921& 0.8920&0.7843&0.9289 &0.9285 &0.8565 &- &- & - \\ 
                    S3&0.9266 &0.9274 &0.8530&0.8861& 0.8828&0.7728&0.9248 &0.9249 &0.8498 &- &- & - \\ 
                             \hline
                    Avg&\underline{0.9262} &\underline{0.9268} &\underline{0.8522}&0.8983 &0.8963 &0.7970 &\bf{0.9309}&\bf{0.9307}&\bf{0.8611}&0.6162 &0.4304 & 0.2623  \\ 

                    \hline
                    \end{tabular}
          
                \label{tab:metrics-J2020}
            \end{table*}

            \begin{table*}[t!]
                \centering
                \caption{Cross-Year Experiment Metrics Comparison: Targeting Jackson County 2021. The best metrics are indicated in bold, while the second-best metrics of the baseline methods are underlined.}
                   
                    \begin{tabular}{|c|ccc|ccc|ccc|ccc|}
                    \hline
                    & \multicolumn{3}{c|}{\textbf{CropSTGAN}}& \multicolumn{3}{c|}{\textbf{CropTGAN}} & \multicolumn{3}{c|}{\textbf{STDAN}} & \multicolumn{3}{c|}{\textbf{TempCNN}} \\ 
                    \hline
                    &OA  & F1  &  Kappa &OA  & F1  &  Kappa  & OA  & F1  &  Kappa & OA   & F1  &  Kappa           \\ 
                    S1&\bf{0.9212} & \bf{0.9219} & \bf{0.8421}&0.8884&0.8814&0.7785&\underline{0.9199} &\underline{0.9197} & \underline{0.8391} &0.5789 &0.3249 & 0.1868 \\ 
                    S2& 0.9182 & 0.9194 & 0.8360              &0.8837&0.8934&0.7653&0.9006 &0.9002 &0.8000 &- &- & - \\ 
                    S3& 0.9198 & 0.9208 & 0.8393              &0.9057&0.8985&0.8124&0.9167 &0.9163 &0.8322 &- &- & - \\ 
                             \hline
                    Avg&\bf{0.9197}&\bf{0.9207}&\bf{0.8391}&0.8926 &0.8911 &0.7854&\underline{0.9124} &\underline{0.9121} &\underline{0.8238} &0.5789 &0.3249 & 0.1868 \\ 

                    \hline
                    \end{tabular}
          
                \label{tab:metrics-J2021}
            \end{table*}

            \begin{figure*}[t!]
                \centering
                \subfloat[]{\includegraphics[width=1.2in]{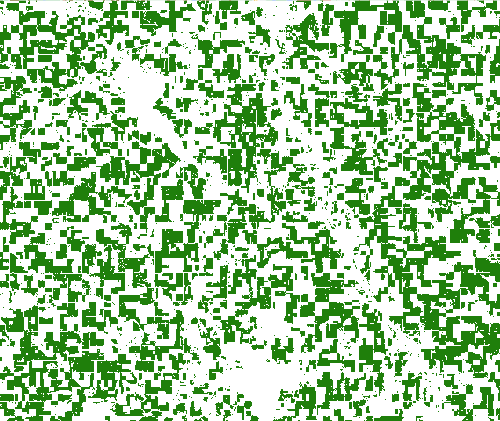}}%
                \label{fig:gt_USA}
                \hfil                
                \subfloat[]{\includegraphics[width=1.2in]{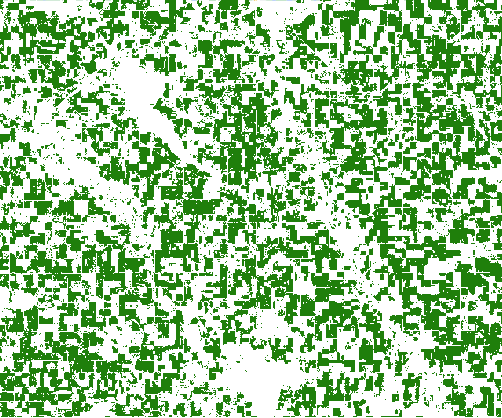}}%
                \label{fig:gt_USA}
                \hfil
                \subfloat[]{\includegraphics[width=1.2in]{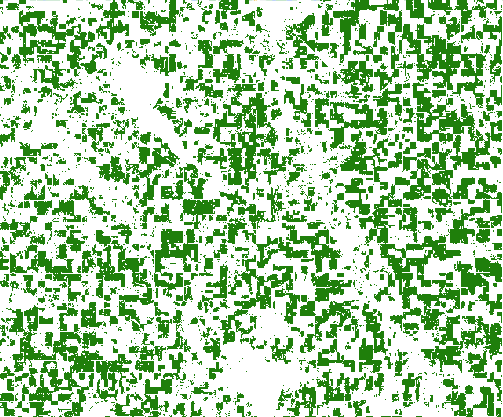}}%
                \label{fig:result_n_China}
                \hfil
                \subfloat[]{\includegraphics[width=1.2in]{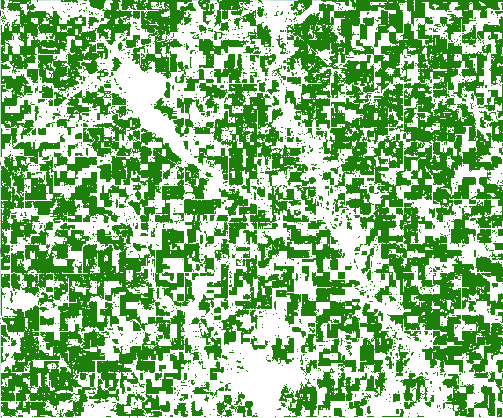}}%
                \label{fig:result_d_China}
                \hfil
                \subfloat[]{\includegraphics[width=1.2in]{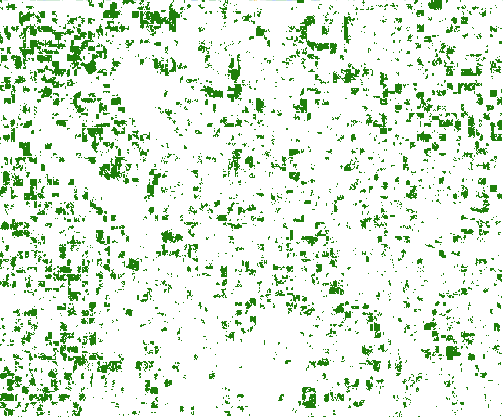}}%
                \label{fig:result_b_China}
                \hfil

                \hspace{1.36in}
                \subfloat[]{\includegraphics[width=1.2in]{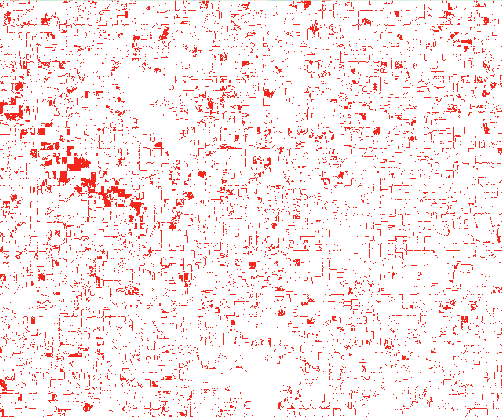}}%
                \label{fig:result_n_China}
                \hfil                
                \subfloat[]{\includegraphics[width=1.2in]{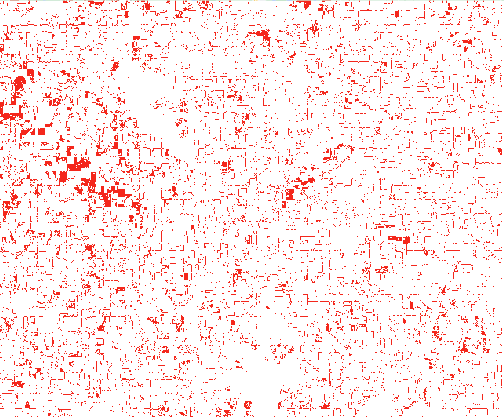}}%
                \label{fig:result_n_China}
                \hfil
                \subfloat[]{\includegraphics[width=1.2in]{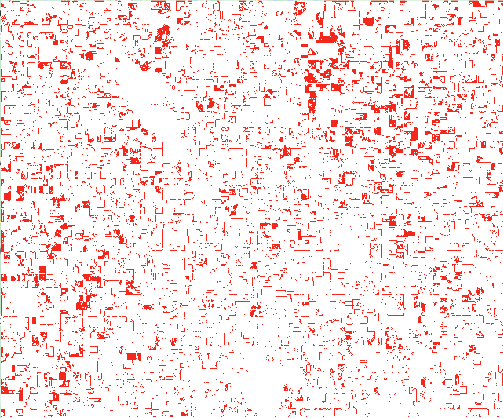}}%
                \label{fig:result_d_China}
                \hfil
                \subfloat[]{\includegraphics[width=1.2in]{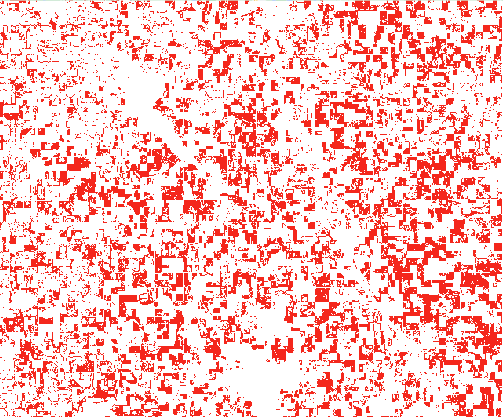}}%
                \label{fig:result_b_China}
                \caption{Corn Crop Mapping Results Comparison with Jackson County 2020 as the Target Domain. (a) displays the ground truth. The crop mapping results are depicted in (b) for CropSTGAN, (c) for CropTGAN, (d) for STDAN, and (e) for TempCNN. In this visualization, green denotes corn, and white represents other types. The corresponding error images are illustrated in panels (f) and (i) for CropSTGAN, CropTGAN, STDAN, and TempCNN, respectively. Red highlights the misclassified pixels, and white represents correctly classified pixels.}
                \label{fig:result_2020}
            \end{figure*}
            \begin{figure*}[t!]
                \centering
                \subfloat[]{\includegraphics[width=1.2in]{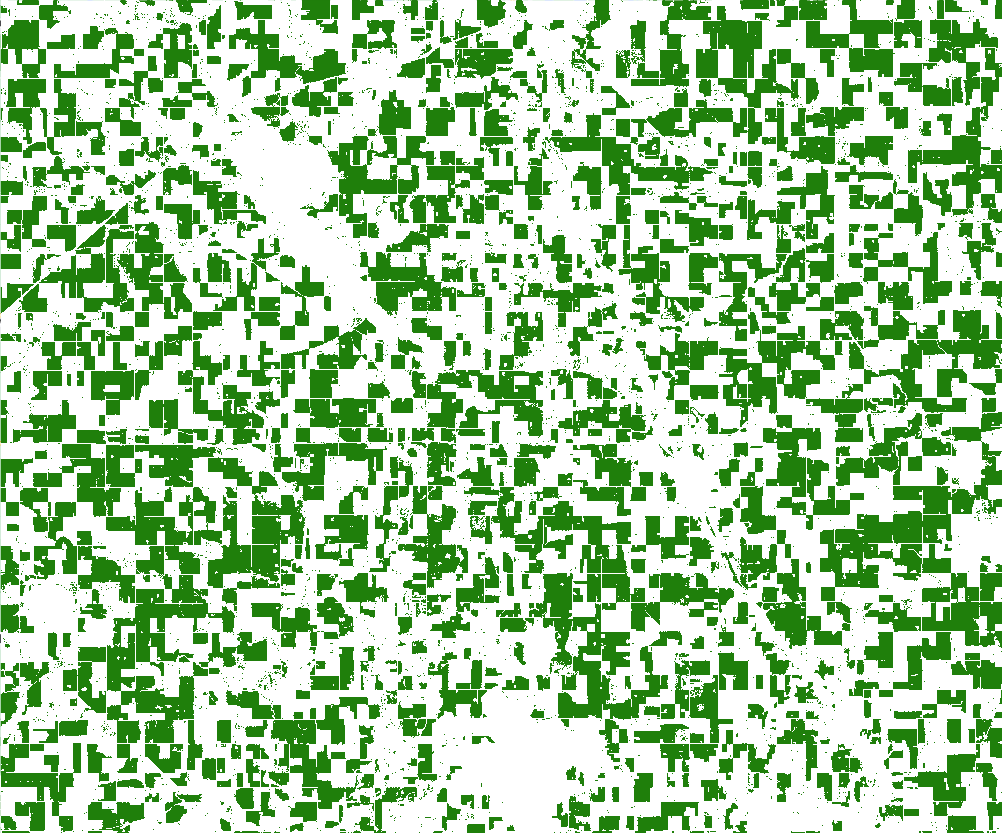}}%
                \label{fig:gt_USA}
                \hfil
                \subfloat[]{\includegraphics[width=1.2in]{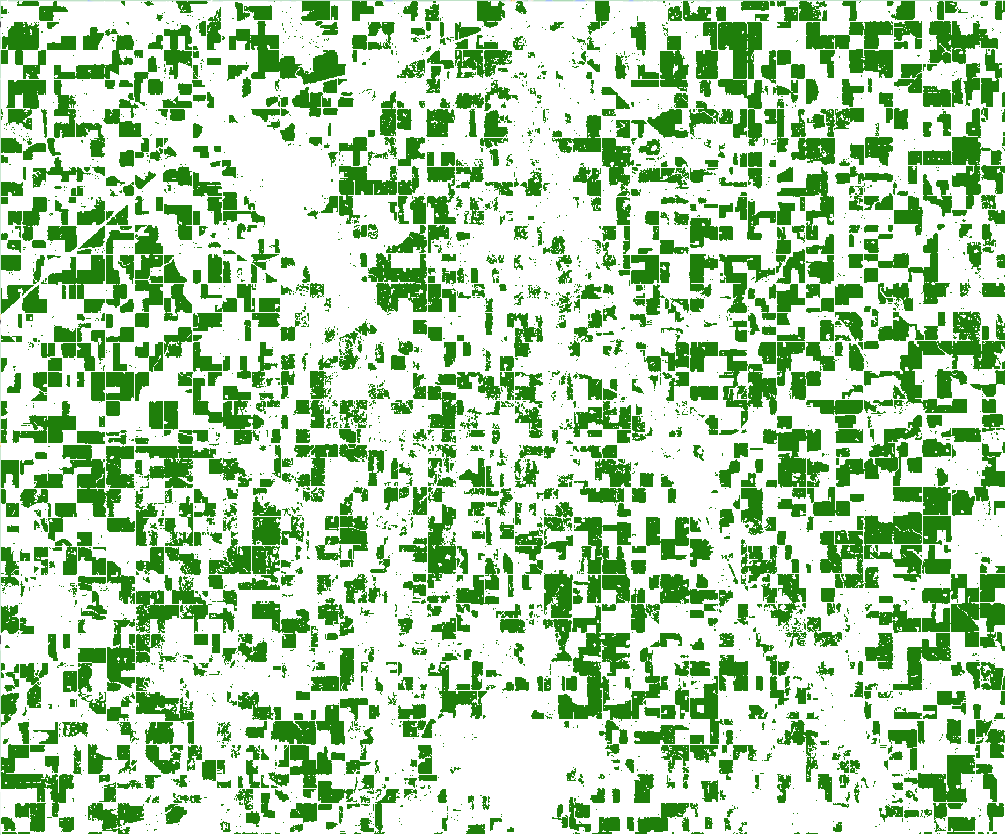}}%
                \label{fig:result_n_China}
                \hfil
                \subfloat[]{\includegraphics[width=1.2in]{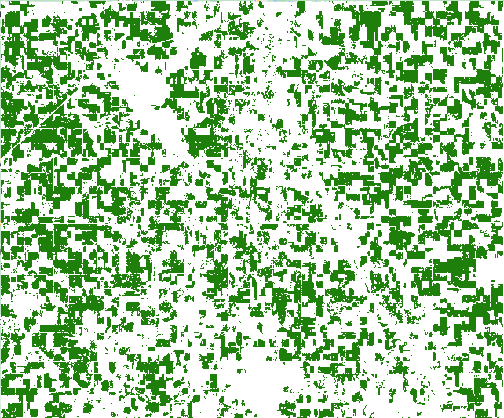}}%
                \label{fig:result_n_China}
                \hfil
                \subfloat[]{\includegraphics[width=1.2in]{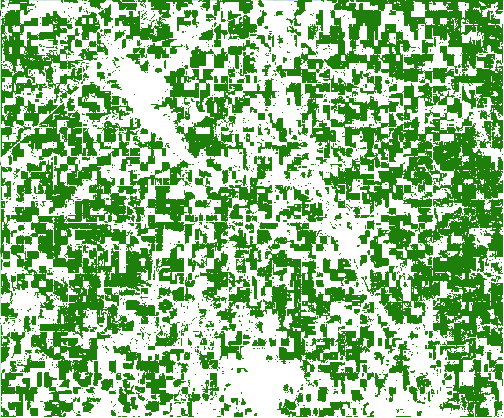}}%
                \label{fig:result_d_China}
                \hfil
                \subfloat[]{\includegraphics[width=1.2in]{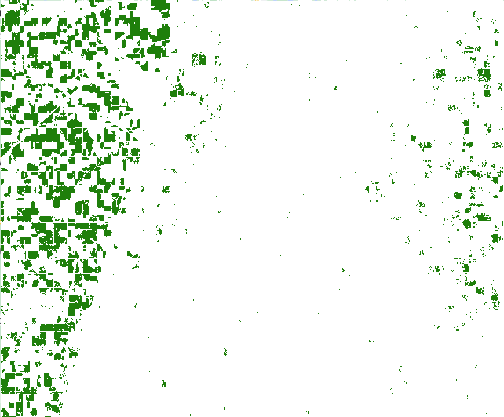}}%
                \label{fig:result_b_China}
                \hfil
                
                \hspace{1.36in}
                \subfloat[]{\includegraphics[width=1.2in]
                {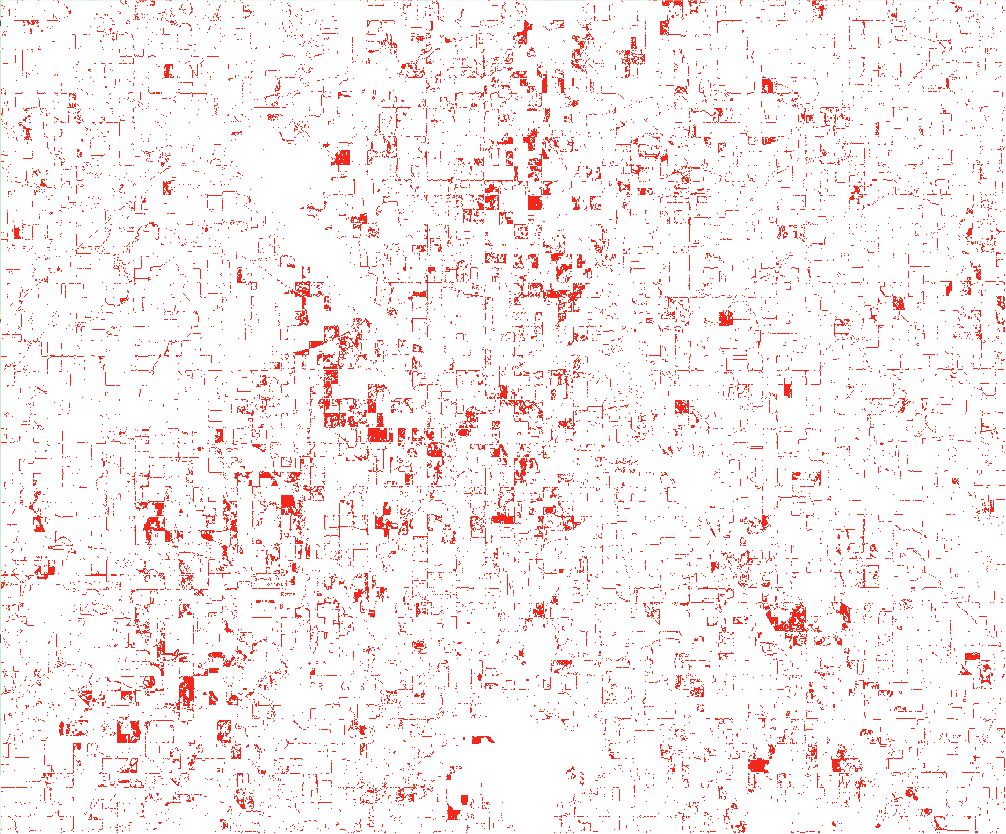}}%
                \label{fig:result_n_China}
                \hfil                
                \subfloat[]{\includegraphics[width=1.2in]{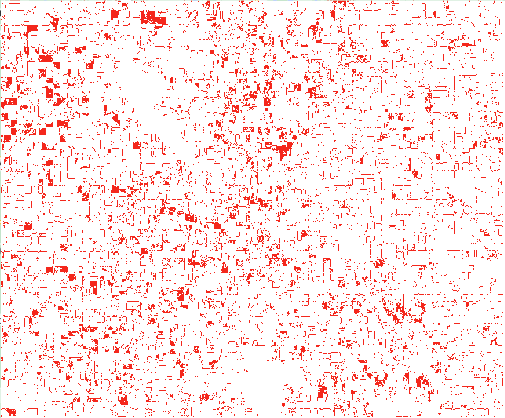}}%
                \label{fig:result_n_China}
                \hfil
                \subfloat[]{\includegraphics[width=1.2in]{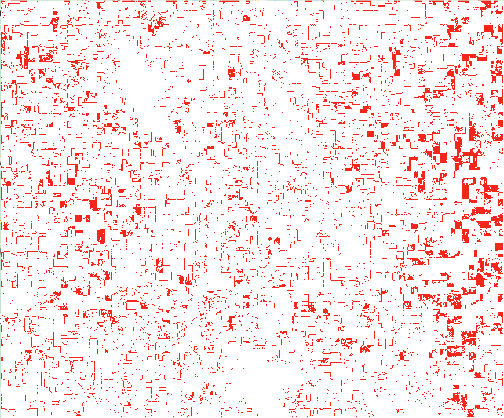}}
                \label{fig:result_d_China}
                \hfil
                \subfloat[]{\includegraphics[width=1.2in]{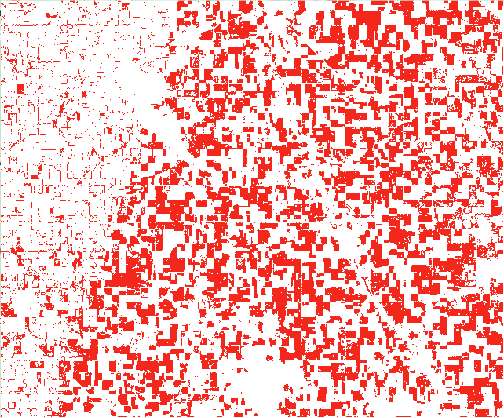}}
                \label{fig:result_b_China}
                \caption{Corn Crop Mapping Results Comparison with Jackson County 2021 as the Target Domain.}
                \label{fig:result_2021}
            \end{figure*}

            \begin{table*}[t!]
                \centering
                \caption{Cross-Region Experiment Metrics Comparison: Targeting China 2019. The best metrics are indicated in bold, while the second-best metrics of the baseline methods are underlined.}
                   
                    \begin{tabular}{|c|ccc|ccc|ccc|ccc|}
                    \hline
                    & \multicolumn{3}{c|}{\textbf{CropSTGAN}} &\multicolumn{3}{c|}{\textbf{CropTGAN}} & \multicolumn{3}{c|}{\textbf{STDAN}} & \multicolumn{3}{c|}{\textbf{TempCNN}} \\ 
                    \hline
                    &OA  & F1  &  Kappa &OA  & F1  &  Kappa  & OA  & F1  &  Kappa & OA   & F1  &  Kappa           \\ 
                    S1&\bf{0.8694} &\bf{0.8762} &\bf{0.7401}&\underline{0.8228} &\underline{0.7961} &\underline{0.6483} & 0.7839 &0.7289 & 0.5622 &0.6268 &0.5232 & 0.2435 \\ 
                    S2 &0.8447 &0.8239 &0.6874              &0.7933 &0.7533 &0.5903 &0.7480 &0.6732 &0.4888 &- &- & - \\ 
                    S3 &0.8331 &0.8061 & 0.6634             &0.7971 &0.7578 &0.5978 &0.7090 &0.6084 &0.4084 &- &- & - \\ 
                             \hline
                    Avg&\bf{0.8491} &\bf{0.8354} &\bf{0.6970}&\underline{0.8044} &\underline{0.7691} &\underline{0.6121}&0.7470 &0.6702 &0.4865 &0.6268 &0.5232 & 0.2435 \\ 

                    \hline
                    \end{tabular}
          
                \label{tab:metrics-china}
            \end{table*}

            \begin{table*}[t!]
                \centering
                \caption{Cross-Region Experiment Metrics Comparison: Targeting Jackson County 2019. The best metrics are indicated in bold, while the second-best metrics of the baseline methods are underlined.}
                   
                    \begin{tabular}{|c|ccc|ccc|ccc|ccc|}
                    \hline
                    & \multicolumn{3}{c|}{\textbf{CropSTGAN}} & \multicolumn{3}{c|}{\textbf{CropTGAN}} & \multicolumn{3}{c|}{\textbf{STDAN}} & \multicolumn{3}{c|}{\textbf{TempCNN}} \\ 
                    \hline
                    &OA  & F1  &  Kappa&OA  & F1  &  Kappa  & OA  & F1  &  Kappa & OA   & F1  &  Kappa           \\ 
                    S1&\bf{0.8303}&0.8181&\bf{0.6630}&0.7881 &0.7899 & 0.5815& 0.7655 & 0.7651 & 0.5286 &0.6662 &0.6905 &  0.3282 \\ 
                    S2&\bf{0.8303}&0.8188&0.6627&0.7806 &\underline{0.8013} &0.5569& 0.7568 &0.7744 & 0.5106 &- &- & - \\ 
                    S3&0.8190&\bf{0.8330}&0.6347&\underline{0.7914} &0.7970 &\underline{0.5821}& 0.7216 &0.6883 &0.4534 &- &- & - \\ 
                             \hline
                    Avg&\bf{0.8265}&\bf{0.8233}&\bf{0.6535}&\underline{0.7867} &\underline{0.7961} &\underline{0.5735}&0.7480 &0.7426 &0.4975  &0.6662 &0.6905 &  0.3282 \\ 

                    \hline
                    \end{tabular}
          
                \label{tab:metrics-china-r}
            \end{table*}
            
            \begin{figure*}[t!]
                \centering
                \subfloat[]{\includegraphics[width=1.2in]{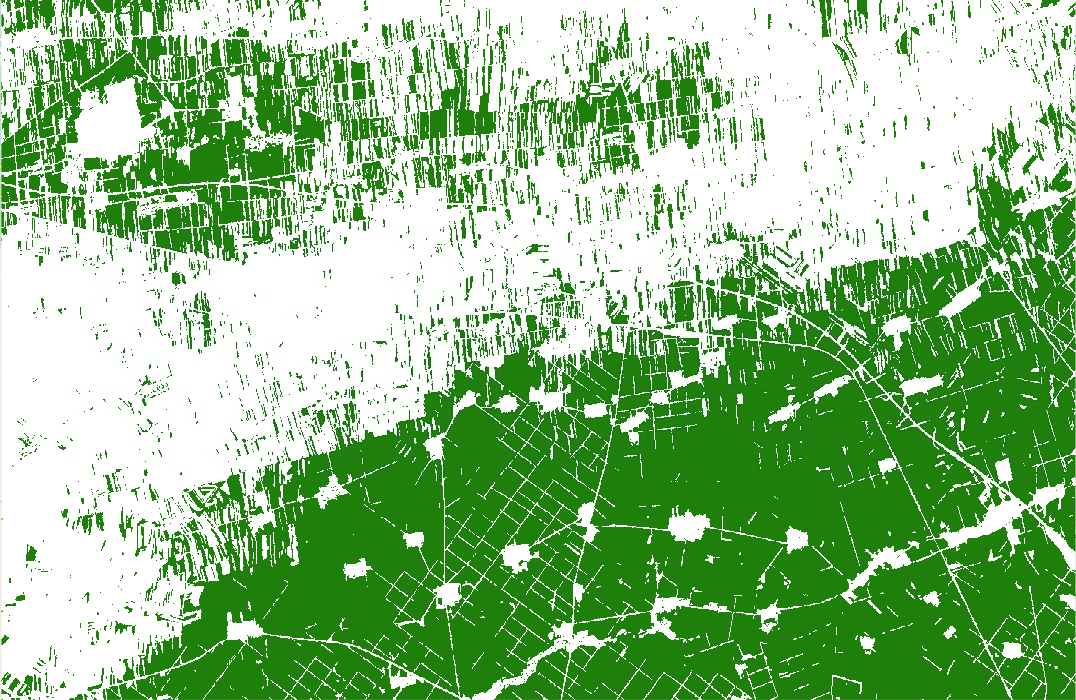}}%
                \label{fig:gt_China}
                \hfil
                \subfloat[]{\includegraphics[width=1.2in]{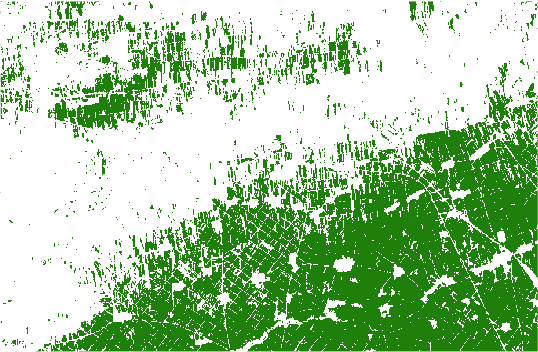}}%
                \label{fig:result_n_China}
                \hfil
                \subfloat[]{\includegraphics[width=1.2in]{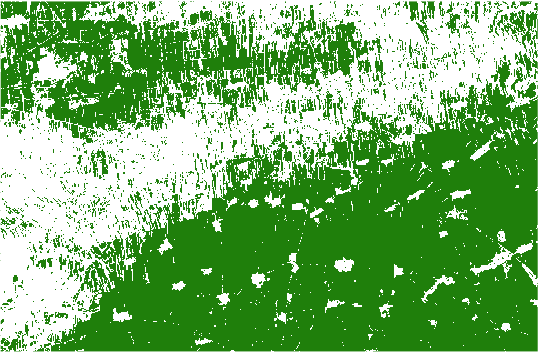}}%
                \label{fig:result_n_China}
                \hfil
                \subfloat[]{\includegraphics[width=1.2in]{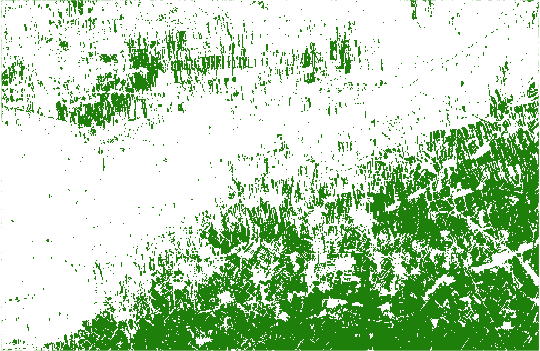}}%
                \label{fig:result_n_China}
                \hfil
                \subfloat[]{\includegraphics[width=1.2in]{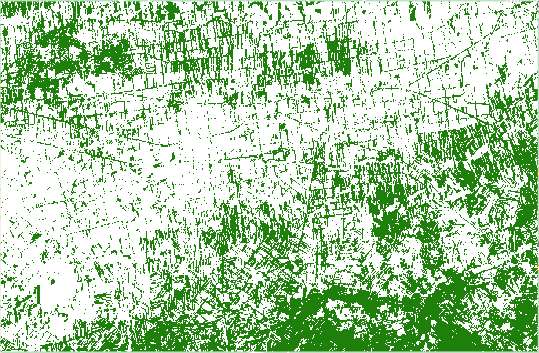}}%
                \label{fig:result_b_China}
                \hfil
                
                \hspace{1.36in}
                \subfloat[]{\includegraphics[width=1.2in]{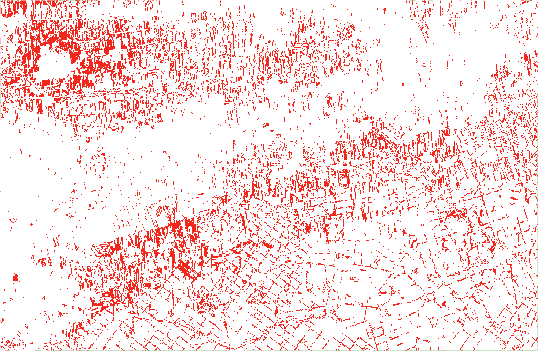}}%
                \label{fig:result_n_China}
                \hfil                
                \subfloat[]{\includegraphics[width=1.2in]{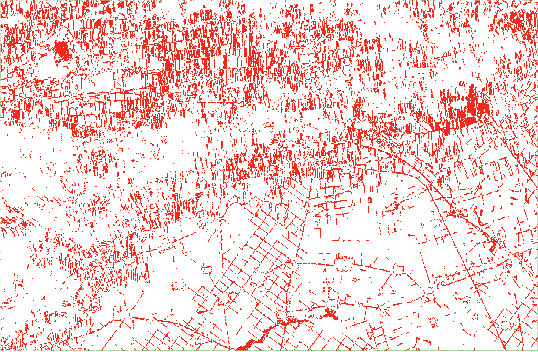}}%
                \label{fig:result_n_China}
                \hfil
                \subfloat[]{\includegraphics[width=1.2in]{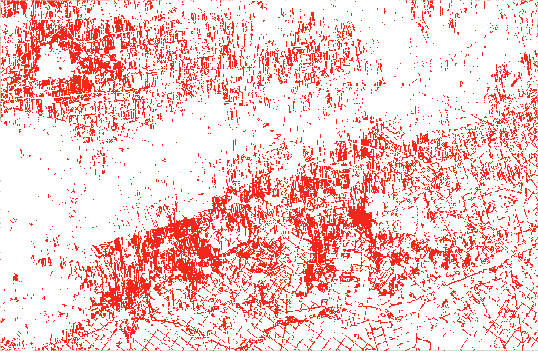}}%
                \label{fig:result_n_China}
                \hfil
                \subfloat[]{\includegraphics[width=1.2in]{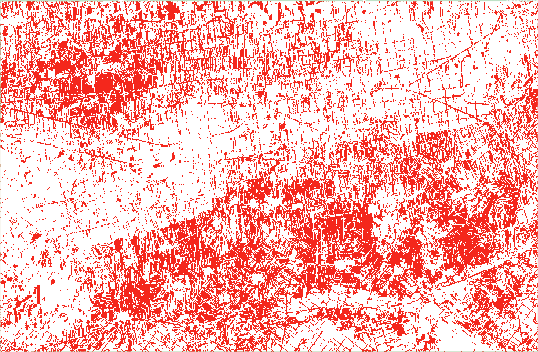}}%
                \label{fig:result_b_China}
                \caption{Corn Crop Mapping Results Comparison with the Study Area of China 2019 as the Target Domain.}
                \label{fig:result_China}
            \end{figure*}
            \begin{figure*}[t!]
                \centering
                \subfloat[]{\includegraphics[width=1.2in]{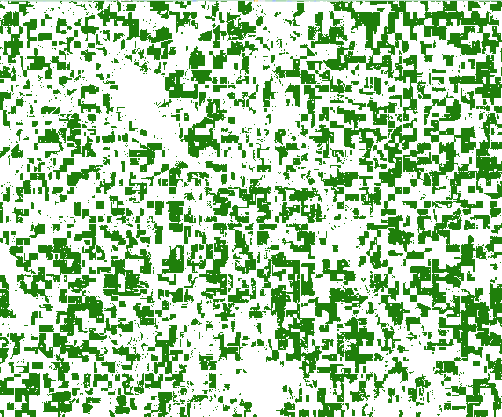}}%
                \label{fig:gt_China}
                \hfil
                \subfloat[]{\includegraphics[width=1.2in]{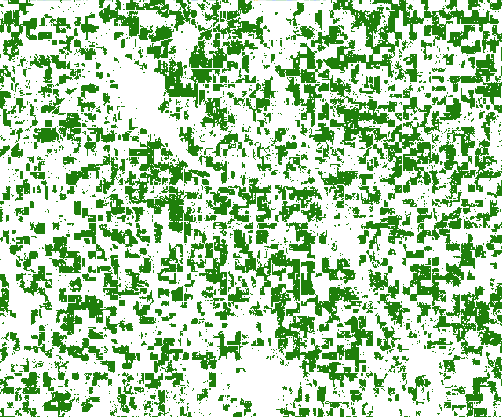}}%
                \label{fig:result_n_China_Temp}
                \hfil
                \subfloat[]{\includegraphics[width=1.2in]{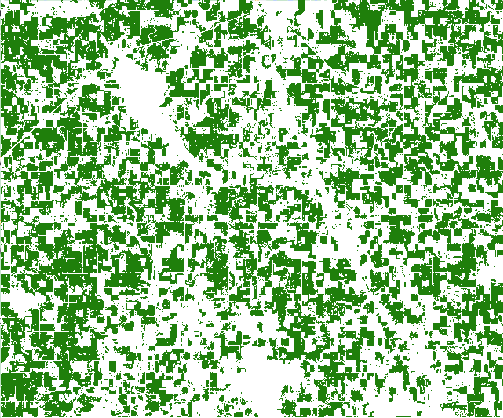}}%
                \label{fig:result_n_China}
                \hfil
                \subfloat[]{\includegraphics[width=1.2in]{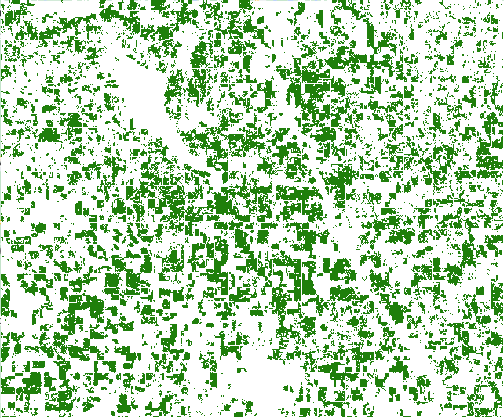}}%
                \label{fig:result_n_China}
                \hfil
                \subfloat[]{\includegraphics[width=1.2in]{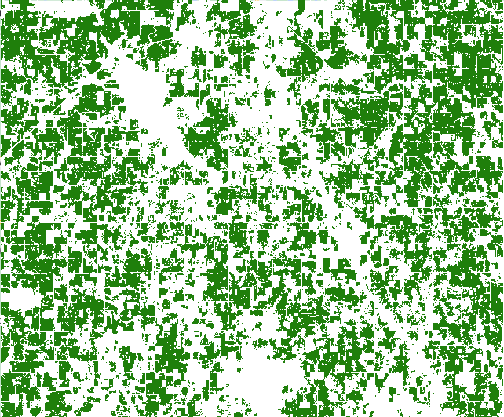}}%
                \label{fig:result_b_China}
                \hfil
                
                \hspace{1.36in}
                \subfloat[]{\includegraphics[width=1.2in]{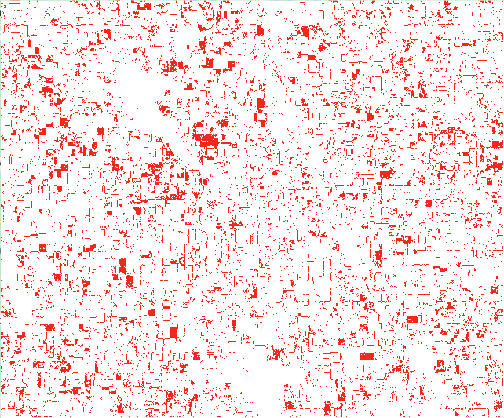}}%
                \label{fig:result_n_China_Temp}
                \hfil
                \subfloat[]{\includegraphics[width=1.2in]{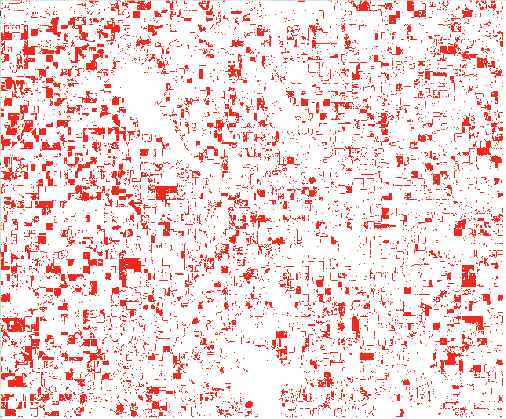}}%
                \label{fig:result_n_China}
                \hfil
                \subfloat[]{\includegraphics[width=1.2in]{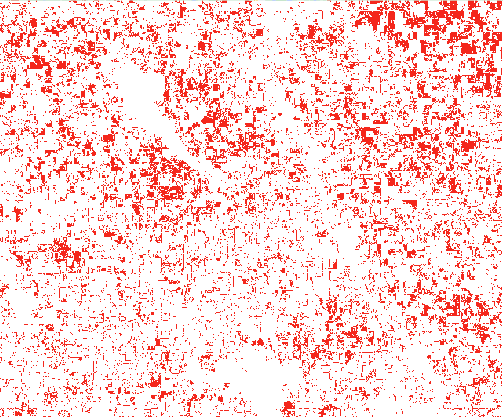}}%
                \label{fig:result_n_China}
                \hfil
                \subfloat[]{\includegraphics[width=1.2in]{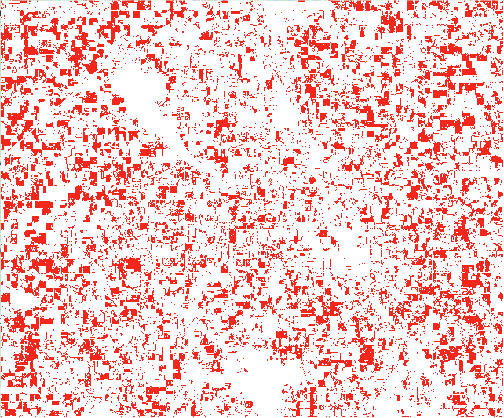}}%
                \label{fig:result_b_China}
                \caption{Corn Crop Mapping Results Comparison with Jackson County 2019 as the Target Domain.}
                \label{fig:result_China_r}
            \end{figure*}

    \subsection{Experiment Setup}
    \label{sec:es}   
        This work designs two sets of experiments to comprehensively evaluate the cross-domain performance of the proposed CropSTGAN method:
        
        \begin{itemize}
            \item The first set of experiments investigates the cross-year scenario, where Jackson County from 2020 to 2021 is considered as the target domain, and Jackson County in 2019 serves as the source domain. The discrepancies in data distribution between these domains are relatively slight.
            \item  The second series of experiments explores cross-region scenarios, incorporating a study area in China and Jackson County. When Jackson County 2019 serves as the source domain, the target domain is the study area in China 2019. Conversely, when the study area in China 2019 is the source domain, Jackson County 2019 becomes the target domain. The discrepancies in data distribution across the domains are relatively large.
        \end{itemize}

        
        
        
        To evaluate the distinct architecture of the CropSTGAN domain mapper, comparative experiments were undertaken utilizing a simpler, analogous structure named CropTGAN. Within this framework, both the discriminators' and generators' encoders share the same filter numbers and sizes as those in CropSTGAN but focus solely on extracting features from the temporal dimension. Additionally, we benchmarked the CropSTGAN framework against various SOTA methods, including TempCNN and STDAN. 
        
        \subsection{Training Setup}
        
        In each experiment, $200,000$ data points are randomly sampled from both the source and target domains, resulting in a total of $400,000$ data points. The TempCNN is trained using $70\%$ of randomly selected labelled data from the source domain. The remaining $30\%$ is divided equally into validation and test datasets. To train the CropSTGAN model, all the sampled data from the target domain and the source domain are used. During training, a batch size of $64$ is used. All Methods' training persisted until the completion of 500 epochs or upon convergence, as determined by an early stopping criterion set at 50 epochs. For optimization, the Adaptive Moment Estimation (Adam) optimizer is employed with a learning rate of $0.005$ and an exponential decay rate of $0.9$ for the first moment estimates. 
        
        Every method was repeatedly trained on each subset from scratch 3 times with the same training configuration. Notably, the TempCNN model was only trained once and served as the crop mapper for the CropSTGAN, resulting in a single TempCNN outcome for each test. Moreover, in the CropSTGAN and CropTGAN framework training processes, it's worth noting that the coefficients of the total loss function in the first set of experiments are set to $\alpha=1, \beta=10, \sigma=5$. For the second set of experiments, they are set to $\alpha=1, \beta=20, \sigma=10$. The choice of coefficients in our total loss function is based on practical experience and is designed to ensure the stability and effectiveness of the CropSTGAN and CropTGAN training processes.

        \subsection{Evaluation Metrics}

        To assess the performance of the binary corn crop mapping, the following metrics are used:

        \subsubsection*{\bf Overall Accuracy (OA)} represents the proportion of all correctly classified items to the total number of items in the dataset. 


        \subsubsection*{\bf F1 Score} is a single metric that combines precision and recall to provide an overall measure of a model's accuracy in classification tasks. It balances the trade-off between correctly identifying positive instances and minimizing false positives and false negatives.

        \subsubsection*{\bf The Kappa coefficient} measures the agreement between the predicted and observed classifications, taking into account the agreement that would occur by chance alone.

    \subsection{Results}
    
    \label{sec:es}     
        \begin{figure*}[]
            \centering
            \subfloat[]{\includegraphics[width=3.2in]{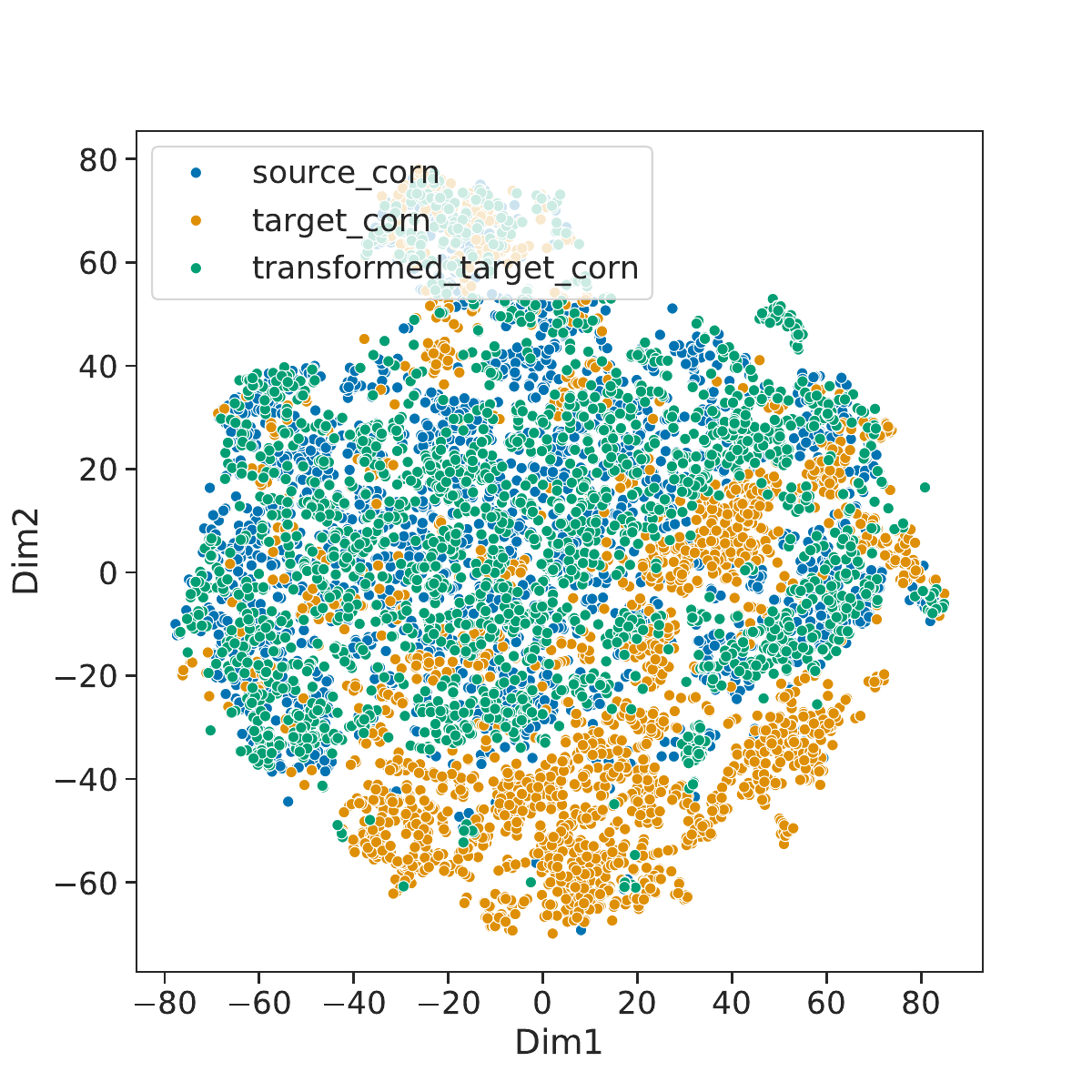}}%
            \label{fig:J2020}
            \hfil
            \subfloat[]{\includegraphics[width=3.2in]{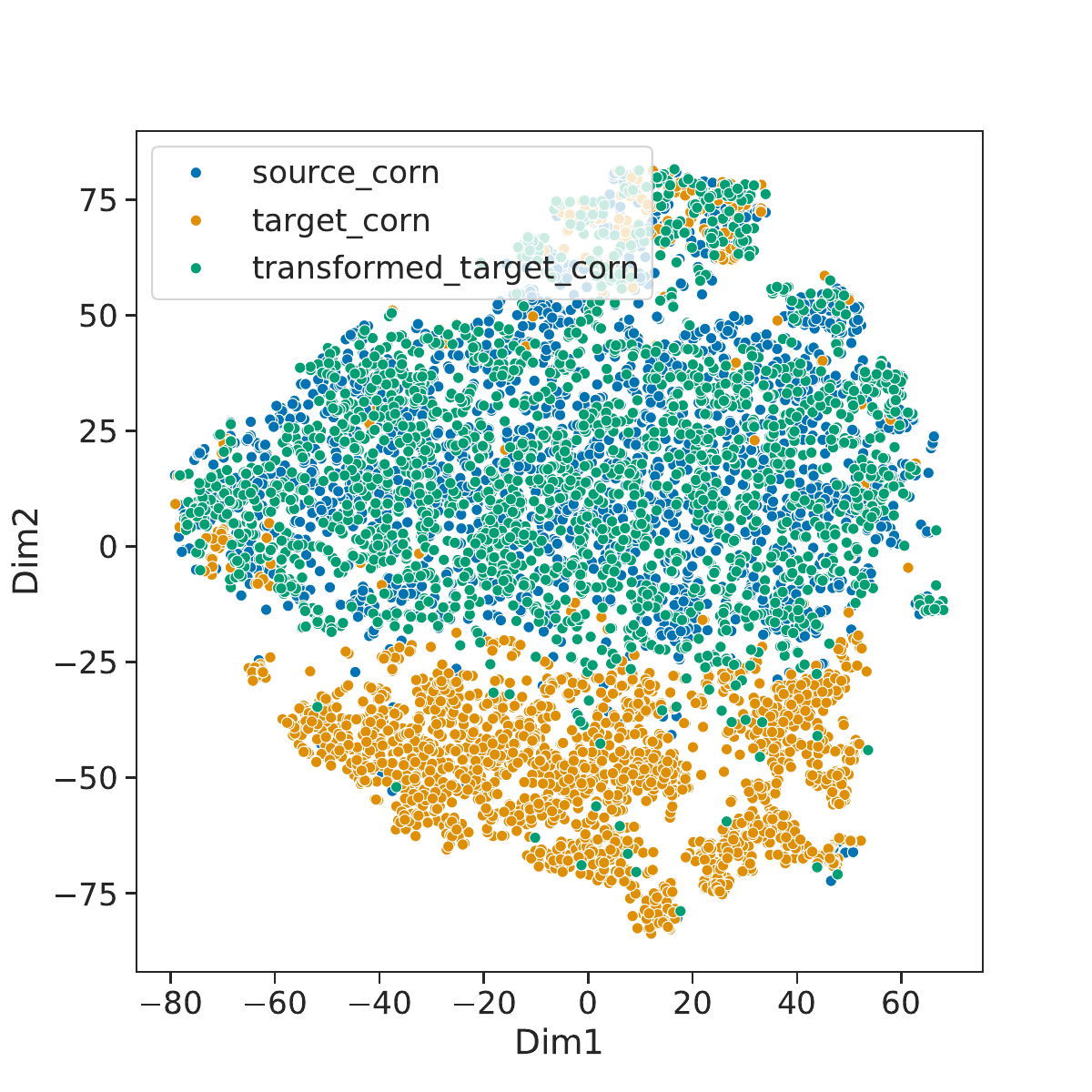}}%
            \label{fig:J2021}

            \caption{The t-SNE Visualization of Corn Data Points for the Cross-Year Experiments: Comparison between Target Domain Data, Transformed Target Domain Data, and Source Domain Data. (a) Jackson County 2020 as the target domain. (b) Jackson County 2021 as the target domain.}
            \label{fig:tsne_cy}
            \end{figure*}

            \begin{figure*}[]
            \centering
            \subfloat[]{\includegraphics[width=3.2in]{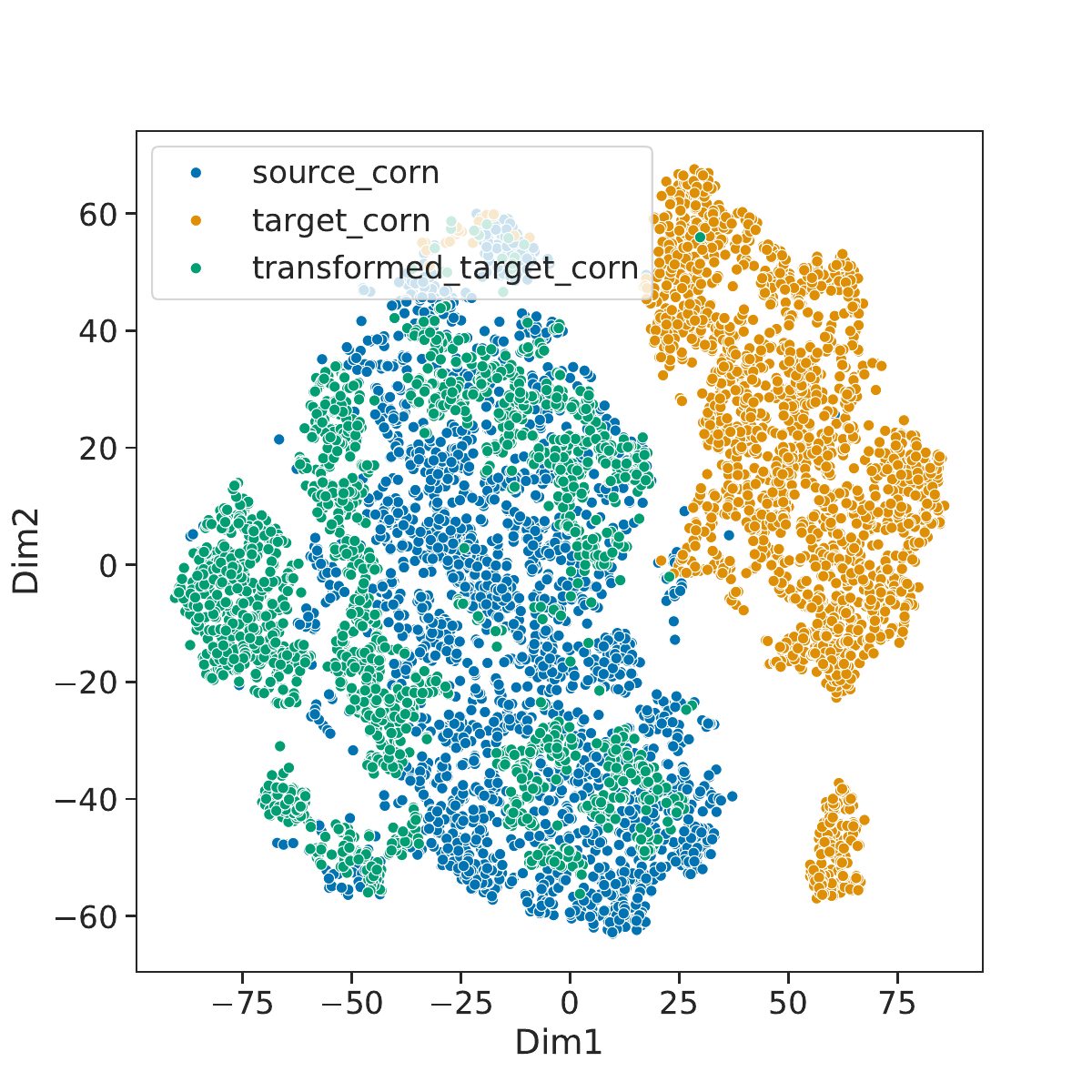}}%
            \label{fig:J2020}
            \hfil
            \subfloat[]{\includegraphics[width=3.2in]{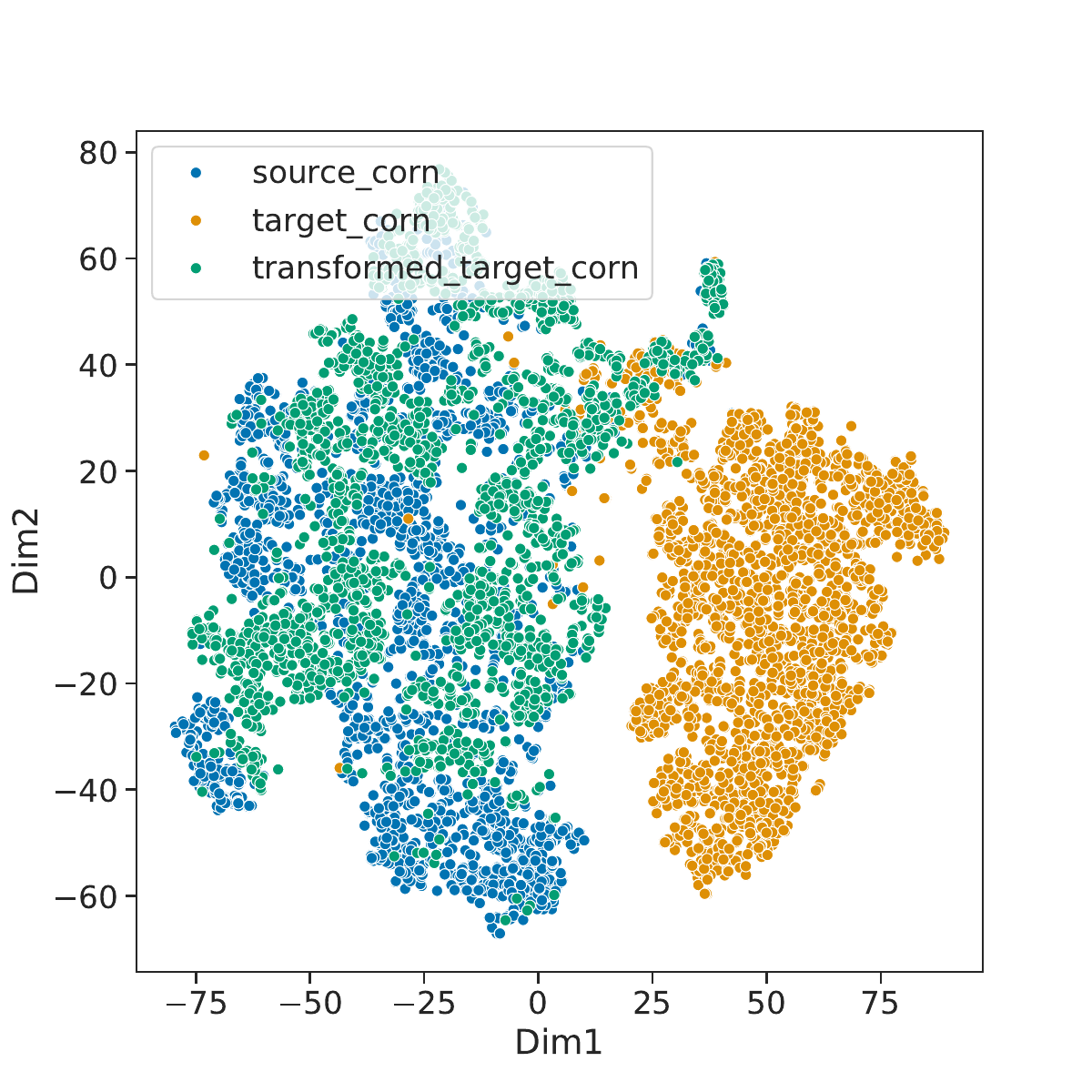}}%
            \label{fig:J2021}

            \caption{The t-SNE Visualization of Corn Data Points for the Cross-Domain Experiments: Comparison between Target Domain Data, Transformed Target Domain Data, and Source Domain Data. (a) The study area of China as the target domain. (b) Jackson County 2019 as the target domain.}
            \label{fig:tsne_cd}
        \end{figure*}

        Tables \ref{tab:metrics-J2020} and \ref{tab:metrics-J2021} present the results for the first set of cross-year experiments. Tables \ref{tab:metrics-china} and \ref{tab:metrics-china-r} present the results for the second set of cross-region experiments. Across two sets of experiments, our CropSTGAN method demonstrates the highest average metrics, except for the cross-year experiment targeting Jackson County 2020.

        In the first set of experiments targeting Jackson County 2020, the CropSTGAN achieved remarkable results, securing second place with an average OA of 92.62\%, an F1 score of 92.68\%, and a Kappa coefficient of 85.22\%. These results represent significant improvements over TempCNN, with increases of +31.00\% in OA, +49.64\% in F1, and +58.99\% in Kappa. Additionally, CropSTGAN outperformed CropTGAN, underscoring the effectiveness of its unique structure. However, it was slightly outpaced by the STDAN, which took first place with an OA of 93.09\%, an F1 score of 93.07\%, and a Kappa of 86.11\%. When Jackson County 2021 plays as the target domain, CropSTGAN outperforms STDAN, leading by a slight margin with an increase of +0.73\% in OA, +0.86\% in F1 Score, and +1.53\% in Kappa Coefficient. Meanwhile, CropTGAN ranks third, significantly surpassing TempCNN in performance. Figures \ref{fig:result_2020} and \ref{fig:result_2021} display the visualization of cross-year experimental results and their error images using CropSTGAN, CropTGAN, STDAN, and TempCNN.

        In the second set of experiments, CropSTGAN outperformed all, securing the top spot, followed by CropTGAN in both tests. Specifically, when targeting the study area of China, CropSTGAN achieved an OA of 84.91\%, an F1 score of 83.54\%, and a Kappa coefficient of 69.70\%. These metrics are markedly higher than those of STDAN, showing increases of 10.21\% in OA, 16.52\% in F1, and 21.05\% in Kappa. Conversely, with China as the source domain and Jackson County 2019 as the target domain, CropSTGAN reached an OA of 82.65\%, an F1 score of 82.33\%, and a Kappa coefficient of 65.35\%, surpassing STDAN by 7.85\% in OA, 8.07\% in F1, and 15.6\% in Kappa. Additionally, CropTGAN, CropGAN, and STDAN all performed better than TempCNN, which only managed an OA of 66.62\%, an F1 score of 69.05\%, and a Kappa coefficient of 32.82\%. Figures \ref{fig:result_China} and \ref{fig:result_China_r} present the results and error images from cross-region experiments.

        \subsection{The t-SNE Visualization}
            
            To assess the effectiveness of our CropSTGAN method in addressing the domain shift problem, t-distributed stochastic neighbour embedding (t-SNE) \cite{van2008visualizing} is utilized for the visualization to analyze the distribution of the target data, transformed target data, and source domain data of the target crop. By using t-SNE, the data points are projected into a two-dimensional space while preserving their local relationships. Figure \ref{fig:tsne_cy} and \ref{fig:tsne_cd} show the t-SNE visualization that illustrates the distribution of corn data points for the cross-year and cross-region experiments.

            In the visualization, the orange points represent MSI data points of corn cropland sourced from the target domain, offering a glimpse into the data distribution within that domain. The green points denote the transformed corn cropland MSI data points from the target domain to the source domain, employing our proposed method. Finally, the blue points indicate the original corn cropland MSI data points extracted from the source domain. Upon analyzing the t-SNE visualization, it is evident that the distribution of corn cropland MSI data points between the source and target domains differs. This disparity highlights the presence of a domain shift, which poses challenges for accurate crop mapping under the target domains. However, the application of our CropSTGAN domain mapper resulted in an improvement in the similarity between the data distribution of the transformed target domain data (green points) and the source domain data (blue points), compared to the similarity between the original target domain data (orange points) and the source domain data (blue points). This resemblance enables the TempCNN crop mapper, trained on the source domain, to effectively process the transformed remote sensing data obtained from the target domain. It coincides with our excellent crop mapping results for these years and this county.

\section{Discussion}
\label{sec:discussion}

\subsection{Analysis of crop mapping results}

In the cross-year experiments, CropSTGAN and STGAN showed similar performance, indicating that CropSTGAN effectively addresses the inter-annual cross-domain challenge, comparable to the state-of-the-art (SOTA) method. In cross-region experiments, CropSTGAN achieved better results than STGAN. As evident from Figure \ref{fig:ndvi_curve}, the NDVI discrepancy is more pronounced in cross-region experiments than in cross-year ones. Under the significant differences in data distribution, CropSTGAN outperforms STGAN, benefiting from the identity loss.

Moreover, CropSTGAN outperformed CropTGAN for all experiments, suggesting the effectiveness of the CropSTGAN domain mapper structure. Additionally, the results from cross-domain methods were significantly better than those obtained by directly applying a CNN-based crop classifier, trained on the source domain, to the target domain.

\subsection{Advantages of CropSTGAN}

Most existing crop mapping related studies rely heavily on a large number of local labelled data for modeling and making predictions, and thus tedious and costly sample collection needs to be carried out extensively and frequently. The sharing of collected labelled samples is an effective way to address the dilemma of ground truth sampling. However, due to the differences in climate conditions across regions and years, trained crop classification models may lose their validity when applied to new domains. Therefore, the CropSTGAN was developed to address the distribution discrepancy existed between the source and target domains, that is, cross-domain issue.

In order to address this issue, most SOTA methods, like STDAN, strive to extract invariant features across target and source domains to tackle cross-domain challenges. However, their effectiveness is often limited by significant differences in data distribution between these domains. In contrast, our CropSTGAN demonstrates superior performance in scenarios with large data distribution disparities, as evidenced by our experimental results.

\subsection{Limitations of CropSTGAN}

However, it is important to acknowledge the limitations of our work. One limitation of our CropSTGAN work is the underlying assumption that the primary crops in the target domain and the source domain are consistent. This assumption, which implies uniformity in the dominant crop types, may not always hold true in practice, particularly when considering diverse agricultural practices across different regions. In future research, addressing this limitation and developing methods that can accommodate variations in primary crop types across domains will be a valuable direction for enhancing the robustness and applicability of our approach.

\section{Conclusions}

In conclusion, we introduced the CropSTGAN framework, which integrates a pre-processor, a domain mapper, and a TempCNN crop mapper, specifically designed to overcome the challenges of cross-domain early crop mapping caused by inter-region and inter-year variations. Notably, the framework is versatile, enabling not only crop mapping but also the classification of various land cover types. The CropSTGAN domain mapper is designed to extract both temporal and spectral features from time-series MSI data, effectively transforming target domain data to the source domain. The TempCNN crop mapper, trained by the labelled source domain data, takes the transformed target domain data as input to locate the target crop for the target domain.

Our comprehensive evaluation, conducted across various regions in the USA and China and spanning different years, demonstrates the CropSTGAN framework's superior performance. It outperforms several baseline and SOTA methods, including TempCNN and STDAN, thereby validating its effectiveness and accuracy in cross-domain early crop mapping scenarios, even with large data distribution disparities between the target domain and source domain.

\bibliography {CD}
\bibliographystyle {IEEEtran}

\begin{IEEEbiography}[{\includegraphics[width=1in,height=1.25in,clip,keepaspectratio]{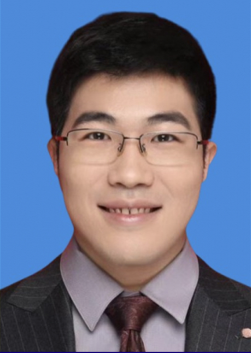}}]{Yiqun WANG} received the M.Sc. degree from the Karlsruhe Institute of Technology, Karlsruhe, Germany, in 2020. He is currently a Ph.D. Candidate with the Interdisciplinary Centre on Security Reliability and Trust, University of Luxembourg, Luxembourg City, Luxembourg. His research interests are in remote sensing, computer vision, and deep learning. 
\end{IEEEbiography}

\begin{IEEEbiography}[{\includegraphics[width=1in,height=1.25in,clip,keepaspectratio]{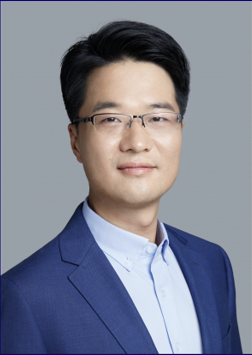}}]{Hui Huang} received the M.Sc. degree in computing science from the University of Glasgow, Glasgow, U.K., in 2013, and the Ph.D. degree from the University of New South Wales, Sydney, NSW, Australia, in 2018. He is currently a Research Associate with the Interdisciplinary Centre on Security Reliability and Trust, University of Luxembourg, Luxembourg City, Luxembourg. His research interests include V2X communications, autonomous driving, and intelligent transportation systems. 
\end{IEEEbiography}


\begin{IEEEbiography}
[{\includegraphics[width=1in,height=1.25in,clip,keepaspectratio]{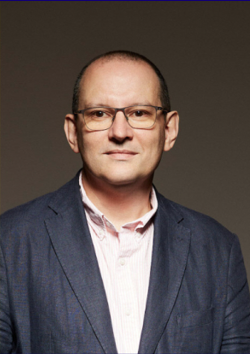}}]{Second B. Author} {Radu STATE} received the M.Sc. degree from the Johns Hopkins University, Baltimore, MD, USA, and the Ph.D. degree and a HDR from the University of Lorraine, Nancy, France. He is a Professor with the Interdisciplinary Center on Security and Trust in Luxembourg. He was a Professor at the University of Lorraine and a Senior Researcher at INRIA Nancy, Grand Est. Having authored more than 100 papers, his research interests cover network and system security and management.  
\end{IEEEbiography}

\EOD

\end{document}